\documentclass[journal]{IEEEtran}
\usepackage{amsmath,amsfonts}
\usepackage{algorithmic}
\usepackage{algorithm}
\usepackage{array}
\usepackage[caption=false,font=normalsize,labelfont=sf,textfont=sf, subrefformat=parens,labelformat=parens]{subfig}
\usepackage{textcomp}
\usepackage{stfloats}
\usepackage{url}
\usepackage{verbatim}
\usepackage{graphicx}
\usepackage[numbers,sort&compress]{natbib}
\usepackage{graphicx} 

\usepackage{amssymb} 

\usepackage{amsthm} 

\newtheorem{theorem}{Theorem}

\newtheorem{definition}{Definition}
\usepackage[breaklinks]{hyperref} 
\hypersetup{
    colorlinks=true, 
    linkcolor=blue, 
    urlcolor=cyan, 
    pdftitle={Categorical Encoding}, 
    citecolor=blue,
}
\usepackage{booktabs} 
\usepackage{tabularx} 
\usepackage{multirow} 
\usepackage[normalem]{ulem} 
\useunder{\uline}{\ul}{} 
\usepackage{threeparttablex} 

\begin{document}

\title{Comparative Study on the Performance of Categorical Variable Encoders in Classification and Regression Tasks}

\author{Wenbin~Zhu,~\IEEEmembership{Member,~IEEE,}
Runwen~Qiu, Ying~Fu
\thanks{
\IEEEcompsocthanksitem W. Zhu and R. Qiu are with the School of Business Administration, South China University of Technology, Guangzhou, Guangdong, 510000, China.\protect\\
E-mail: i@zhuwb.com, qiurunwen@gmail.com.
\IEEEcompsocthanksitem Y. Fu is with the School of Business Administration, South China University of Technology, Guangzhou, Guangdong, China, and the Department of Industrial and Systems Engineering, University of Wisconsin-Madison, Madison, WI, 53706, USA. \protect\\
				E-mail: ying.fu@wisc.edu

}
}



\maketitle

\begin{abstract}
Categorical variables often appear in datasets for classification and regression tasks, and they need to be encoded into numerical values before training. Since many encoders have been developed and can significantly impact performance, choosing the appropriate encoder for a task becomes a time-consuming yet important practical issue. This study broadly classifies machine learning models into three categories: 1) ATI models that implicitly perform affine transformations on inputs, such as multi-layer perceptron neural network; 2) Tree-based models that are based on decision trees, such as random forest; and 3) the rest, such as kNN. Theoretically, we prove that the one-hot encoder is the best choice for ATI models in the sense that it can mimic any other encoders by learning suitable weights from the data. We also explain why the target encoder and its variants are the most suitable encoders for tree-based models.
This study conducted comprehensive computational experiments to evaluate 14 encoders, including one-hot and target encoders, along with eight common machine-learning models on 28 datasets. The computational results agree with our theoretical analysis. The findings in this study shed light on how to select the suitable encoder for data scientists in fields such as fraud detection, disease diagnosis, etc. 
\end{abstract}


\section{Introduction}\label{sec:intro}
\IEEEPARstart{C}{lassification}
and regression finds wide applications across various domains, such as disease diagnosis, product recommendation, demand forecasting, etc. In such tasks, categorical variables like gender, education, city, and occupation are common. Many machine learning models cannot process categorical variables directly, necessitating the use of encoders to convert them into numerical representations. Numerous encoders have been developed in the past, posing a new challenge for model builders: determining the most appropriate category encoder for their specific model or application. To address this challenge, this study aims to provide reliable guidelines via comprehensive numeric experiments and attempt to explain the reasons behind the suitability of specific classes of encoders for certain types of models.

A categorical variable such as ``season" typically includes four distinct \textit{levels}: ``spring", ``summer", ``autumn", and ``winter". The number of levels is called \textit{cardinality}. Categorical variables may convey up to four types of information. Firstly, they inherently partition samples into \textbf{groups}, where all samples within the same level form a group. Secondly, some categorical variables exhibit a natural \textbf{order} among their levels -for example, the progression from ``spring" to ``summer" - which is helpful when the response variables demonstrate a degree of monotonicity with respect to this order. Thirdly, the labels assigned to levels can carry\textbf{ semantic information}. For instance, in the variable ``occupation", ``mechanic technician" shares greater similarity with ``senior engineer technician" than with  ``police aide" \cite{cerda2022encoding}. Lastly, each level defines a conditional distribution of a target variable $y$. Comparing the similarity of conditional distribution allows for measuring the proximity between levels; we refer to this kind of information as \textbf{target information}.
 
To capture the aforementioned information, various encoding methods are available. For extracting grouping information, encoders establish a one-to-one correspondence between levels and encoded values. Examples include the one-hot encoder, contrast encoders \cite{anonym2011library}, and binary encoder \cite{seger2018investigation}. For encoding ordinal information, methods like ordinal encoder and count/frequency encoder \cite{prokopev2018mean} reflect the inherent order among levels. Regarding semantic information, existing methods mainly compute the string similarity. For instance, \citet{cerda2018similarity} computes the string overlap between each pair of levels to derive similarity, while \citet{weinberger2009feature} as well as \citet{cerda2022encoding} indirectly assess string similarity through hash function collision. Regarding target information, the most commonly used technique is the Empirical-Bayes-based target encoder. It utilizes the conditional expectation or conditional probability of the corresponding target at a given level as the encoding.\cite{james1992estimation, micci-barreca2001preprocessing, morris1983parametric, mougan2021quantile, romeijn2017stein, zhou2015shrinkage}.

Given the focus of this article on encoder performance comparison, it is essential to offer a brief overview of previous efforts in the field of encoder analysis. Early research primarily involved evaluating encoder performance within specific scenarios characterized by limited models, datasets, and encoders. \citet{alkharusi2012categorical} studied two contrast encoders in a regression task using a linear regression model. The study found that while their performance was similar, differences in the interpretation of intercepts and coefficients were observed. Similarly, 
\citet{moeyersoms2015including} compared five encoders using a linear support vector machine model for church prediction in the energy supplier domain. The result showed that the best encoder varies under different performance measures. \citet{potdar2017comparative} applied a neural network to a car evaluation dataset to assess seven different encoders. The finding indicated that the contrast encoder yielded the highest accuracy among the evaluated encoders. \citet{seger2018investigation} focused on three encoding methods: one-hot, binary, and feature hashing. Logistic regression and neural networks were employed to conduct experiments on two datasets. The research findings revealed that the one-hot encoding method demonstrated the best performance. \citet{wright2019splitting} studied how to handle categorical variables in random forests. The results on nine datasets showed that target encoder was the most effective method. In the survey paper, \citet{hancock2020survey} discussed five general encoders and introduced various embedding techniques tailored to specific scenarios. Furthermore, several studies compared encoder performance while proposing new models and encoders, offering valuable insights for encoder selection \cite{cerda2018similarity, prokhorenkova2018catboost, johnson2021encoding, mougan2021quantile, cerda2022encoding}.

Recent comparative studies obtained empirical rules through extensive comparative experiments involving multiple datasets, models, and encoders.
\citet{valdez-valenzuela2021measuring} conducted experiments on five different models across ten datasets, comprising seven natural datasets and three synthetic datasets. The primary finding of this research suggested that the CatBoost encoder demonstrated the most favorable overall performance, while one-hot encoder showed notable efficacy in logistic regression. \citet{seca2021benchmark} tested 16 encoding methods across 15 regression datasets using seven distinct predictive models. The research indicated that the top-performing methods were the target encoder and its variants. In the comparative research conducted by \citet{pargent2022regularized}, nine encoders and five machine learning models were employed across 24 datasets that include high-cardinality variables. The result indicated that the regularized version of target encoder performs well on high-cardinality features. Additionally, the target encoder combining the Generalized Linear Mixed Model (GLMM) performed the best.

In conclusion, early research provided valuable insights within specific contexts, but the scale of these studies made it challenging to establish a general rule for encoder selection. Recent comparative studies, while including the latest models and encoders, primarily focused on the average performance of encoders. However, there is a gap in theoretical analysis regarding the connections between models and encoders, as well as between datasets and encoders. Therefore, this work aims to address this gap by theoretically analyzing the applicable conditions of encoders and providing general guidance based on numerous comparative experiments.

The rest of the paper is structured as follows: Section~\ref{sec:onehot} provides proof demonstrating that, given sufficient data, the one-hot encoder can reproduce any other encoders for models executing affine transformation on input, such as logistic regression and multi-layer perceptron neural network. Section~\ref{sec:target} delves into the analysis of the reasons and applicable conditions leading to the good performance of target encoder on tree-based models like Random Forest and Gradient Boosting Decision Trees). Our extensive experiments and analyses are presented in Section~\ref{sec:experiment}. Section~\ref{sec:guide} offers a comprehensive summary and guide for encoder selection and Section~\ref{sec:conclusion} concludes this work. 
 
\section{One-hot as universal encoder for ATI} \label{sec:onehot}

\subsection{Reproducing other encoders} \label{subsec:onehot_repro}

For a categorical variable $x$ has $c$ levels, i.e., distinct values, denoted as $\mathcal{V}=\{v_1,v_2,\dots,v_c\}$, an encoder is a function $\phi: \mathcal{V} \rightarrow \mathbb{R}^l$ that maps each level $v \in \mathcal{V}$ into a $l$-dimensional vector $\phi(v)$. Taking the categorical variable ``season" as an example: the one-hot encoder maps ``spring", ``summer", ``autumn", and ``winter" into four-dimensional vectors $\left[1,0,0,0\right]^T, \left[0,1,0,0\right]^T, \left[0,0,1,0\right]^T$, and $\left[0,0,0,1\right]^T,$ respectively. Alternatively, the ordinal encoder maps each season into 1-dimensional number: 1, 2, 3, and 4, respectively.

We call several well-known classification and regression models \emph{ATI models} due to the affine transformation they applied to input variables. Among these, neural networks stand as typical examples. Consider the multi-layer perceptron neural network (MLPNN), where the input (before applying the activation function) of the first hidden layer, denoted as $\boldsymbol{z}$, is an affine transformation of input vector $\boldsymbol{x}$. More precisely, $\boldsymbol{z}$ is the product of a weight matrix $\boldsymbol{W}$ and the input vector $\boldsymbol{x}$ plus a bias term $\boldsymbol{b}$. Other ATI models include linear regression (LNR), logistic regression (LGR), and support vector machine (SVM) with linear kernel.

\begin{figure}[htbp]
	\centering
        \subfloat[Generic Encoder]{
		\includegraphics[width=0.47\linewidth]{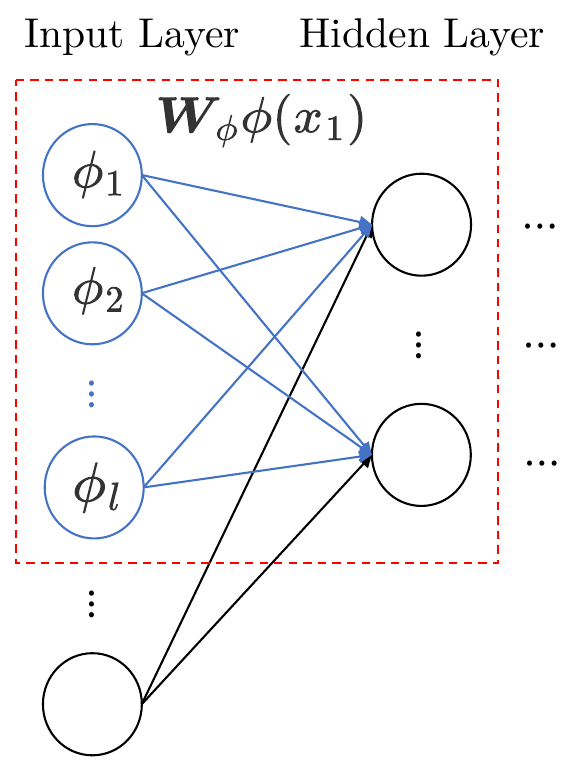}
		\label{fig:sketch_mlp_general}
        }
        \subfloat[One-hot Encoder]{
		\includegraphics[width=0.47\linewidth]{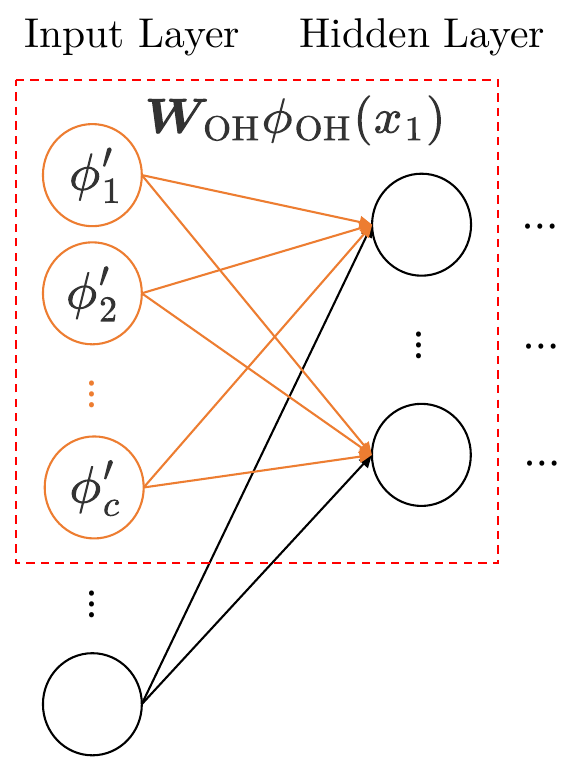}
		\label{fig:sketch_mlp_onehot}
        }
	\caption{(a) A multi-layer neural network. A categorical variable, denoted as $x_1$ with a cardinality of $c$, is encoded by a generic encoder $\phi(x_1) = [\phi_1, \cdots, \phi_{l}]^T$ and the contribution of $x_1$ to input of first hidden layer is given by blue sub-network $\boldsymbol{z}_\phi = \boldsymbol{W}_\phi \phi(x_1) $ (b) Encoder is replaced by one-hot $\phi_\text{OH}(x_1) = [\phi'_1, \cdots, \phi'_{c}]^T$ and the contribution of $x_1$ to input of first hidden layer is given by the orange sub-network.}
	\label{fig:infer_temp_when_summer}
\end{figure}

Figure~\subref*{fig:sketch_mlp_general} illustrates a neural network model applied to inputs containing categorical variables. Without loss of generality, we arrange input variables so that all categorical variables precede other variables, as their order does not influence the training process. While we illustrate our idea with a single categorical variable, it can be easily extended to multiple categorical variables. Upon applying the encoder $\phi(\cdot)$, the input vector becomes $\left[\phi(x_1) \quad \boldsymbol{x}_o\right]^T$, where $\boldsymbol{x}_o$ corresponds to all other non-categorical variables. The weight matrix $\boldsymbol{W} = \left[\boldsymbol{W}_\phi \quad \boldsymbol{W}_o\right]$ of the first layer are partitioned accordingly where $\boldsymbol{W}_\phi$ corresponds to encoded values $\phi(x_1)$. The input (before applying the activation function) to the first layer, denoted as $\boldsymbol{z}$, is decomposed into $\boldsymbol{z}_\phi$—representing the contribution due to the categorical variable $x_1$—and $\boldsymbol{z}_o$, denoting the contribution from other variables and the bias term.
\begin{eqnarray}
	\boldsymbol{z} &=& \boldsymbol{W} \left[\phi(x_1) \quad \boldsymbol{x}_o\right]^T + \boldsymbol{b} \nonumber \\ 
	           &=& \boldsymbol{W}_\phi \phi(x_1)  + \left(\boldsymbol{W}_o \boldsymbol{x}_o + \boldsymbol{b}\right) \nonumber \\ 
	           &=& \boldsymbol{z}_\phi + \boldsymbol{z}_o .
\end{eqnarray}
Figure~\subref*{fig:sketch_mlp_onehot} illustrate the same NN where encoder $\phi$ is replaced by one-hot encoder $\phi_\text{OH}$. Let $\boldsymbol{W}_\text{OH}$ be the weight sub-matrix corresponding to $\phi_\text{OH}(x_1)$. The contribution to input of first layer, $\boldsymbol{z}_\phi$, due to categorical variable $x_1$ can be reproduced by $\phi_\text{OH}(x_1)$ if the following holds,
\begin{equation}\label{eqn:ati}
	\boldsymbol{z}_\phi = \boldsymbol{W}_\text{OH} \phi_\text{OH}(x_1)  \quad \forall x_1 .
\end{equation}
When equation \eqref{eqn:ati} holds, the neural network trained with one-hot encoder $\phi_\text{OH}$ will produce exactly the same output as neural network trained with any generic encoder $\phi$ for any input. The generic encoder $\phi$ can be any hand-crafted encoders such as ordinal encoder, target encoder. 

For any given weight $\boldsymbol{W}_\phi$ and training data, we can solve the following system of linear equations to obtain $\boldsymbol{W}_\text{OH}$ that satisfies equation \eqref{eqn:ati}
\begin{eqnarray}
	\boldsymbol{W}_\text{OH} \phi_\text{OH}(x_1) = \boldsymbol{W}_\phi \phi(x_1) \quad \forall x_1 .
\end{eqnarray}
It is not hard to show by linear algebra that the above equation always has a solution
\begin{equation*}
\boldsymbol{W}_\text{OH}=\left[\boldsymbol{W}_\phi \phi(v_1) \quad \boldsymbol{W}_\phi \phi(v_2) \quad \cdots \quad \boldsymbol{W}_\phi \phi(v_c)\right].
\end{equation*} 
We summarize our main result as the following theorem,

\begin{theorem}\label{theorem:onehot_mimic}
	For a categorical variable $\mathcal{V}=\{v_1,v_2,\dots,v_c\}$ and any encoder $\phi: \mathcal{V} \rightarrow \mathbb{R}^l$ that encodes any level $x \in \mathcal{V}$ into $l$ dimension vector $\phi(x)$, an arbitrary linear transformation of $\phi(x)$, $f(x) = \boldsymbol{W}_\phi\phi(x)$, there exists an equivalent linear transformation $h(x) = \boldsymbol{W}_\text{OH}\phi_{\text{OH}}(x)$ on values encoded by one-hot $\phi_{\text{OH}}(x)$, i.e. 
    \begin{equation*}\label{equ:onehot_mimic_in_afftran}
        h(x)=f(x) \quad\forall x \in \mathcal{V} .
    \end{equation*}
\end{theorem}

\subsection{Convergence behavior}


Although the one-hot encoder, when equipped with suitable weights $\boldsymbol{W}_\text{OH}$, can reproduce the outcomes produced by any other encoder, these weights are practically learned from training data. Therefore, sufficient data is essential to learn such weights. Our investigation reveals that \emph{the average number of samples per level (ASPL)} serves as an excellent indicator for data sufficiency. This makes sense since we need sufficient data to estimate the distribution of the target variable for each level. In this section, we illustrate how ASPL influences the performance gap between the one-hot encoder and the best encoder for ATI models using synthetic datasets. Additionally, in Section~\ref{subsec:perf_of_ecd_in_ati}, we delve into this influence on natural datasets.

Suppose target $y$ is generated by a categorical variable $x$ and a Gaussian noise $\epsilon$ as 
\begin{equation}\label{equ:syn_data1}
	y = \phi(x) + \epsilon ,
\end{equation}
where $x\in\{\text{``spring", ``summer", ``autumn", ``winter"}\}$ follows a uniform distribution, and $\phi$ maps these seasons into 0, 1, 2, and 3, respectively. The exact mapping $\phi$ is unknown to the modeler but is considered the best encoder. To assess the performance, a random test set with 1000 samples is generated. We vary the ASPL values within $\{5, 10, 15, \cdots 100\}$ with an interval of 5. For each ASPL value, we randomly generate 30 training sets with different random seeds. For each training set, the categorical variable is encoded by one-hot. Subsequently, a single hidden layer neural network with 100 neurons and ReLU as the activation function is trained and its prediction accuracy is measured by Mean Square Error (MSE) on the test set. The average MSE across the test datasets is computed to evaluate the performance of one-hot encoding at that particular ASPL. 

\begin{figure}[htbp]
    \centering
    \subfloat[Single-Layer NN]{\includegraphics[width=0.48\linewidth]{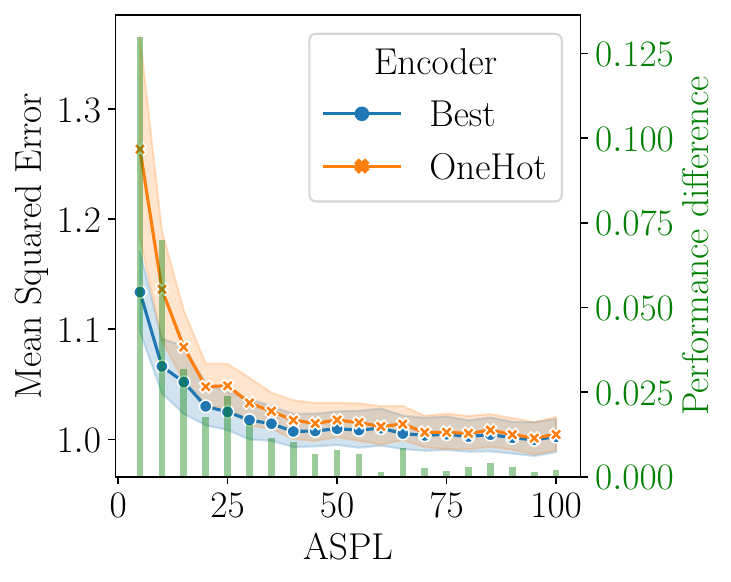}\label{fig:perf_diff_betw_ecd_MLPRegressor}
    }
    \subfloat[Linear Regression]{\includegraphics[width=0.48\linewidth]{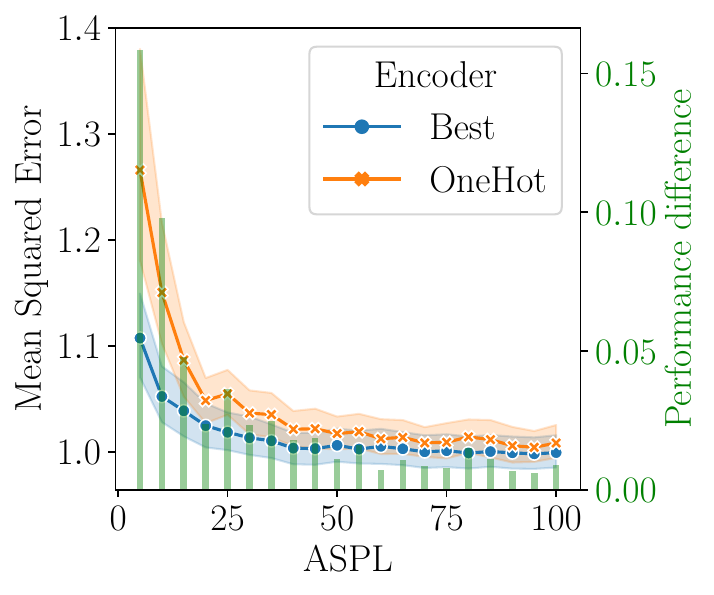}\label{fig:perf_diff_betw_ecd_LinearRegression}
    }
    \caption{Model performance using the best encoder and one-hot encoder on datasets generated from Equation~\eqref{equ:syn_data1}. The x-axis represents the ASPL value.  The left y-axis displays the performance (i.e. Mean Squared Error) using line plots with markers. The right y-axis exhibits the performance difference between two encoders using a bar plot. The shaded region corresponds to the 95\% confidence interval. (a). Single-Layer NN; (b). Linear Regression.} 
    \label{fig:perf_diff_betw_ecd}
\end{figure}

The orange line in Figure~\subref*{fig:perf_diff_betw_ecd_MLPRegressor} illustrates the performance of the Single-Layer Neural Network (NN) utilizing the one-hot encoder. The blue line shows the performance of the best encoder used to generate the training/test datasets. The trend indicates that as ASPL increases, the average performance of the one-hot encoder converges towards the best encoder. The performance difference between them also converges to zero, as indicated in the right y-axis. This observation aligns with our theoretical finding that the one-hot encoder can effectively reproduce any encoder given sufficient data. A similar trend is observed in the Linear Regression model, as shown in Figure~\subref*{fig:perf_diff_betw_ecd_LinearRegression}.

In addition, we try to examine the contribution difference to the first layer of categorical variables using both the best encoder and  the one-hot encoder. More precisely, we look at the vector corresponding to the categorical variables after encoding and linear transformation. The contribution corresponding to the best encoder and the one-hot encoding are denoted as $\boldsymbol{z}_\phi$ and $\boldsymbol{z}_{\text{OH}}$, respectively. We employ the Euclidean distance between the two vectors to quantify their difference. In other words, the \textit{average contribution difference} is computed in the test set $\mathcal{X}_{\text{test}}$ as,
\begin{align*}
    & \quad \frac{1}{|\mathcal{X}_{\text{test}}|}\sum_{x\in \mathcal{X}_{\text{test}}} \|\boldsymbol{z}_\phi -  \boldsymbol{z}_{\text{OH}}\|_2^2 \\
    & = \frac{1}{|\mathcal{X}_{\text{test}}|}\sum_{x\in \mathcal{X}_{\text{test}}}\left \| \boldsymbol{W}_\phi \phi(x) -\boldsymbol{W}_\text{OH}\phi_\text{OH}(x)  \right \|_2^2 .
\end{align*}

\begin{figure}[htbp]
    \centering
        \subfloat[Single-Layer NN]{
        \includegraphics[width=0.48\linewidth]{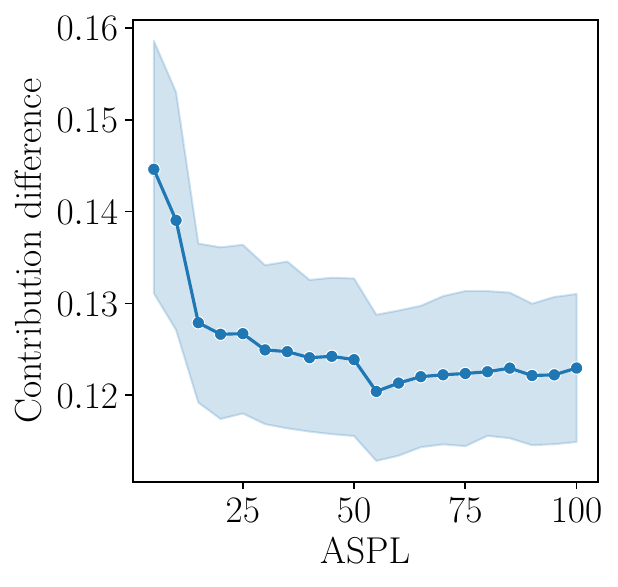}        \label{fig:contr_diff_betw_ecd_MLPRegressor}
    }
    \subfloat[Linear Regression]{
    \includegraphics[width=0.48\linewidth]{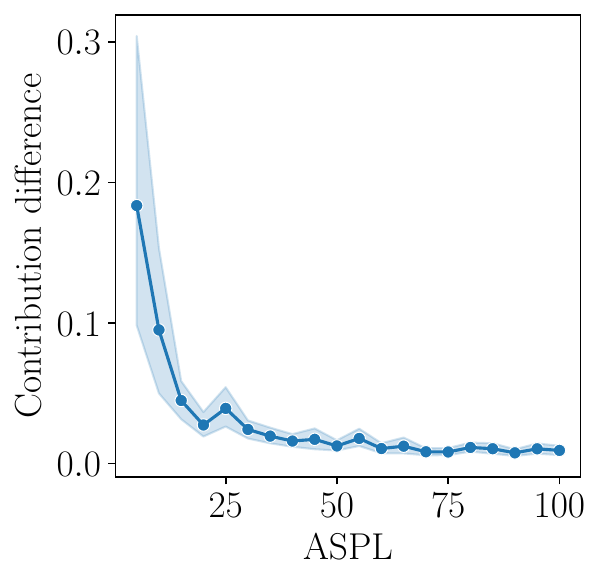}        \label{fig:contr_diff_betw_ecd_LinearRegression}
    }
    \caption{The average contribution difference between the one-hot encoding and the best encoder. The shaded region corresponds to a 95\% confidence interval.(a). Single-Layer NN; (b). Linear Regression. }
    \label{fig:contr_diff_betw_ecd}
\end{figure}

Figure~\subref*{fig:contr_diff_betw_ecd_LinearRegression} illustrates the average contribution difference between the one-hot and the best encoder in the context of Linear Regression model. We can observe that as the ASPL increases, the average contribution difference drops quickly. Additionally, it can gradually approaches zero. Similar contribution difference decrease trend can be observed in Figure~\subref*{fig:contr_diff_betw_ecd_MLPRegressor}, which is in the context of Single-Layer NN. A similar contribution difference decreasing trend can be observed as the ASPL increases. However, the contribution difference remains constant, hovering around 0.12-0.13, regardless of the increase in ASPL. A likely cause is that a complex neural network may have multiple local optima with comparable total loss but different parameter configurations. Consequently, the optimal parameters selected for one-hot may be different from that of the best encoder at the end of the training procedure, leading to differing contributions between one-hot encoding and the best encoder. In general, this behavior is consistent with the prediction of our theory.

\section{Target encoders for tree-based models} \label{sec:target}

\subsection{Optimal split and target encoders}


Encoding high cardinality categorical variables into numerical values can greatly reduces computational burden in tree-based models. The central step in building a binary decision tree is repeatedly splitting a node by choosing the optimal variable that minimize the impurity of the best split. The best split by a given variable is found by exhaustively enumerating all possible splits. For a categorical variable with $c$ levels, there are $(2^c-2)/2$ possible splits -- each level can appear in either left or right child, resulting in total $2^c$ possible splits, subtracting two useless splits where either child is empty and considering only half due to symmetry. Since the number of possible splits grow exponentially with respect to the cardinality $c$, enumerating all splits become intractable. In contrast, when splitting a numerical variable, the tree-based models will evaluate every the midpoint between consecutive unique values as a possible cutoff point between the left and right child. Therefore, a numerical variable with $c$ unique values only offers $c-1$ possible splits. Hence, encoding categorical levels into numerical representations can significantly mitigates computational burden.

The efficiency of computational improvement due to encoding does come with a cost -- encoding a categorical variable usually results in worse splits, since some of the splits are lost. For example, if we encode ``spring", ``summer", ``autumn" and ``winter" into 1, 2, 3, and 4, respectively, there will be only three possible splits after the encoding. These splits correspond to the left child containing either \{``spring"\} (i.e., 1.5 as cutoff), \{``spring", ``summer"\} (i.e., 2.5 as cutoff), or \{``spring", ``summer", ``autumn"\} (i.e., 3.5 as cutoff). It is impossible to put ``spring" and ``autumn" in left child. If the optimal split with minimal impurity involves placing both ``spring" and ``autumn" in the left child, this best split may become unattainable after encoding, potentially leading to a strictly inferior split.

However, it is interesting that \emph{Mean Encoder} and its variants may achieve near-optimal splits under certain conditions\cite{fisher1958grouping}. Consequently, these encoders, grouped together as \textit{Target Encoders}, often demonstrate exceptional performance, particularly for tree-based models among all encoders.

\begin{definition} \emph{Mean Encoder} $\phi_M$ maps a level $v \in \mathcal{V}=\{v_1,v_2,\dots,v_c\}$ of a categorical variable $x$ to the average of target variable $y$ over all training data where $x = v$, which is the simplest estimator of the conditional mean $\mathbb{E}[y \mid x=v]$. \footnote{In binary classification tasks, the encoding corresponds to the conditional probability. Therefore, when the target variable is preprocessed to only have 0s and 1s, the encoding formula used in binary classification tasks aligns with that in regression tasks.} 
\end{definition}

For example, consider a dataset consisting of quarterly sales reports from some stores. In this dataset, the target variable $y$ represents the sales volume, while a categorical variable $x$ represents the season. The Mean Encoder maps ``spring" to the average value of $y$ across all sales records when $x = \text{``spring"}$. 

The finding of \citet{fisher1958grouping} indicated that Mean Encoder will preserve the optimal split at root node when impurity is measured by mean squared error (MSE), a metric suitable for regression tasks. According to Fisher, we may re-index levels in ascending order of their value by the Mean Encoder $\phi_M$,
\begin{equation}
	\phi_M(v_1) \le \phi_M(v_2) \le \cdots \le \phi_M(v_c) ,
\end{equation}
and there exists an optimal split where left child containing the first several levels and the right child containing the remaining levels. This type of split is called contiguous partition by \citet{fisher1958grouping}. \citet{breiman1984classification} proved a similar result: Mean Encoder will preserve the optimal split at root node when impurity is measured by entropy, a suitable metric for classification tasks. To properly apply the results by Fisher and Breiman et. al. to split an internal node, the Mean Encoder $\phi_M$ must be recomputed using the data in that internal node. For example, when a root node is split into left and right child, the average of $y$ where $x=\text{``spring"}$ in left child may be different from that in the root node. The difference can vary significantly. In extreme cases, all data associated with ``spring" may go into the right child, leading to a substantial difference. Conversely, the difference could be very small if the root node is split by an variable that is independent of season and the conditional distribution of $y$ on each level is nearly identical in left and right child.

What if we do not recompute Mean Encoder $\phi_M$ for each internal node, and use only the Mean Encoder $\phi_M$ computed for root node? We argue that in some cases, we may still achieve good results as follows. Since shallow nodes with smaller depth usually contains more data than deeper nodes at larger depth, the split of shallow nodes usually has greater impact than deeper nodes. When there are many numerical variables independent of a categorical variable $x$, it is very likely that the split of shallow nodes are due to those numerical variables, as a result, the conditional distribution of $y$ on levels of $x$ in shallow nodes may be very similar to the root node. In this case, even if we do not recompute Mean Encoder $\phi_M$, we may still obtain good splits. According to the experiments by \citet{wright2019splitting}, applying Mean Encoder at root node of Random Forest may achieve similar performance as trying exhaustive partitioning at each internal node for both binary classification and regression tasks.

\subsection{Performance and data sufficiency}

Target encoders essentially estimates the target variable $y$ conditioned on each level of the categorical variable $x$. Such estimation is accurate only if there are sufficient number of samples per level in the training set. When the number of samples per level is small, the estimated conditional mean may deviates greatly from the ground truth and mislead the partitioning process discussed in the previous section. When there are many levels with small number of samples, the chance of producing wrong partitioning increases and performance of target encoder are expected to be poor. 

We employ a synthetic example in this section to demonstrate the decline in the performance of target encoder as data adequacy diminishes. We measure data sufficiency using the average number of samples per level (ASPL) within the training set. Moreover, a similar trend can be observed in natural datasets, as shown in Section~\ref{subsec:perf_of_ecd_in_tb}.

\begin{figure}[htbp]
	\centering
        \subfloat[]{
		\includegraphics[width=0.47\linewidth]{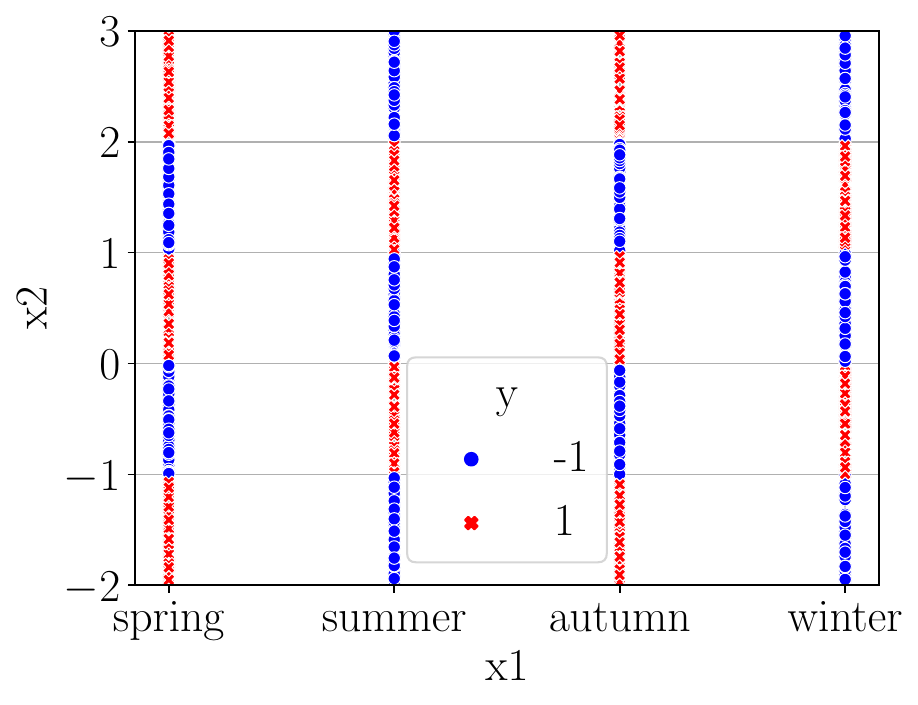}
		\label{fig:binary_cross_testset}
        }
        \subfloat[]{
	\includegraphics[width=0.47\linewidth]{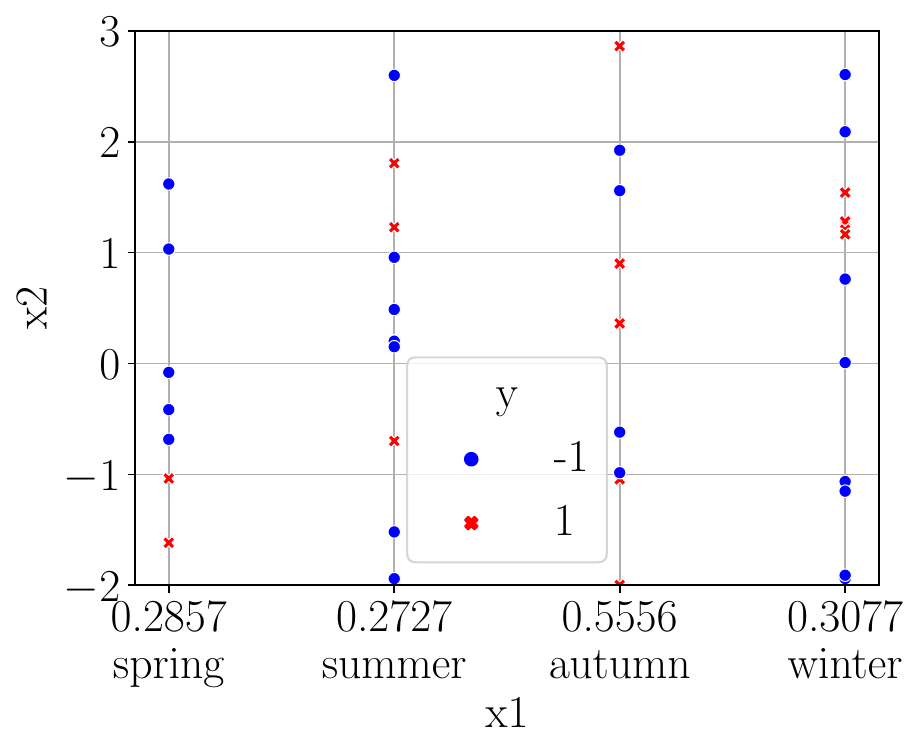}
	\label{fig:binary_cross_mean}
        }
        
        \subfloat[]{
		\includegraphics[width=0.47\linewidth]{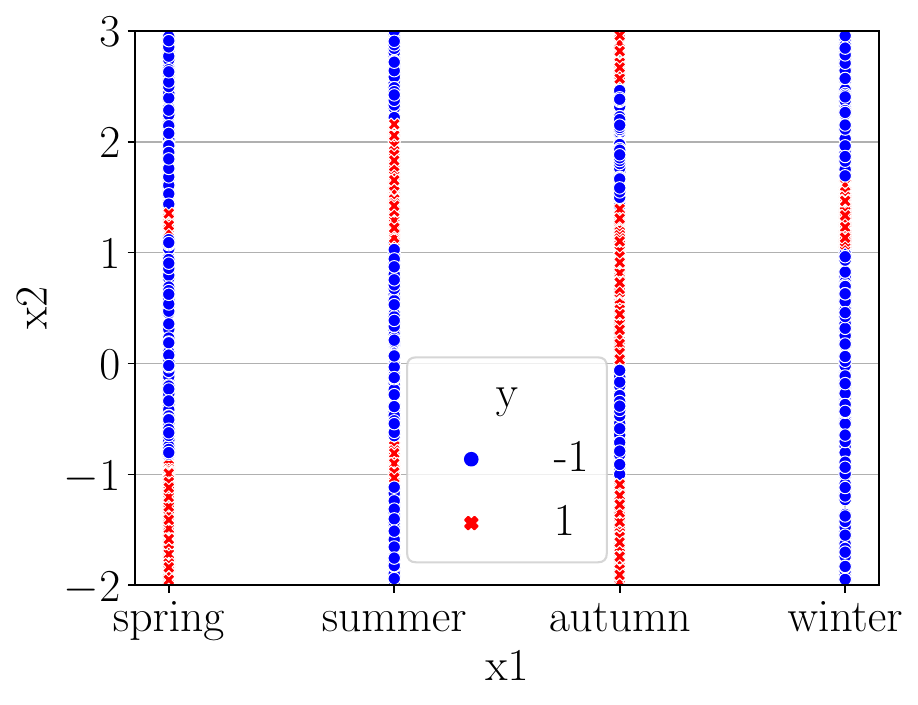}
		\label{fig:binary_cross_mean_pred}
        }
	\caption{(a) Test set with 1000 samples generated using  Equation~\eqref{equ:syn_data2}; (b)  Training set consisting of 40 samples with an ASPL value of 10.  The numerical values above each level are obtained by the Mean Encoder; (c) Prediction on the test dataset by a Random Forest model trained with the aforementioned training set and Mean Encoder, yielding an overall accuracy of 0.730.}
	\label{fig:binary_cross_small_aspl}
\end{figure}

In the synthetic example, target variable $y$ is generated by the following equation:
\begin{equation}\label{equ:syn_data2}
	y=\phi(x_1) \cdot \operatorname{sgn}\left(\sin\left(x_2\pi\right)\right) ,
\end{equation}
where $x_1$ is a uniformly distributed categorical variable ``season" with four levels ``spring", ``summer", ``autumn" and ``winter". The encoder $\phi$ maps these four seasons into 1, -1, 1, and -1 respectively. This encoding scheme is considered the optimal possible encoder but unknown to the machine learning modeler. The sign function $\operatorname{sgn}(\cdot)$ converts positive numbers to 1 and non-positive numbers to -1. The numerical variable $x_2$ is uniformly distributed within the range $[-2, 3]$ and is independent of $x_1$. A total of 1000 samples are generated for the test set using the Equation~\eqref{equ:syn_data2} (see Figure~\subref*{fig:binary_cross_testset}). Simultaneously, a training set consisting of 40 samples is also generated, corresponding to an ASPL value of 10 (see Figure~\subref*{fig:binary_cross_mean}). As we can see, the conditional distribution of $y$ for each level in the training set deviates significantly from the ground truth: 0.2857 for ``spring", 0.2727 for ``summer", 0.5556 for ``autumn", and 0.3077 for ``winter", as opposed to the ground truth values of 0.6 for ``spring", 0.4 for ``summer", 0.6 for ``autumn", and 0.4 for ``winter". Consequently, a Random Forest model trained on the training set with the Mean Encoder exhibits poor performance on the test set, achieving an overall accuracy of approximately 0.730. The corresponding predictions are illustrated in Figure~\subref*{fig:binary_cross_mean_pred}.

Next, we investigate the impact of different ASPL values on the performance of the Mean Encoder. For each ASPL in $\{5,10,15, \cdots, 100\}$ (with an interval of 5), 30 training sets are randomly generated by the Equation~\eqref{equ:syn_data2} with different seeds. A Random Forest model is then trained on each of these training sets using the Mean Encoder, and its performance on the test dataset is measured in terms of accuracy. The average performance of the Mean Encoder for the specified ASPL is computed by averaging the results obtained from the 30 models.

\begin{figure}[htbp]
	\centering
	\includegraphics[width=0.8\linewidth]{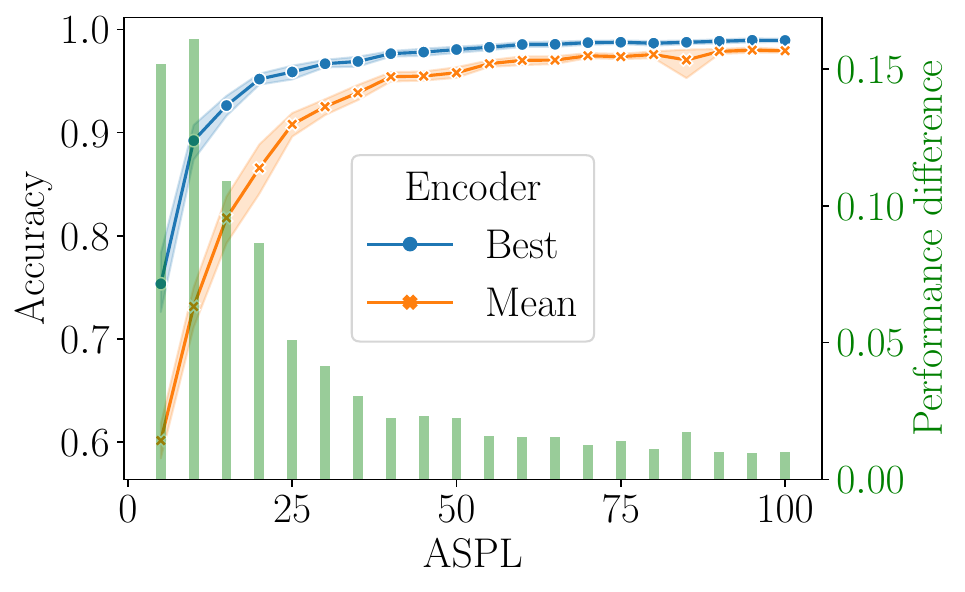}
    \caption{
    Performance of Random Forest using the best encoder and Mean Encoder on datasets generated from Equation~\eqref{equ:syn_data2}. The x-axis represents the ASPL value. The left y-axis displays the performance (i.e., Accuracy) using line plots with markers. The right y-axis exhibits the performance difference between two encoders using a bar plot. The shaded region corresponds to the 95\% confidence interval.}\label{fig:aspl_perf_tree_ecd_relavant_RandomForest}
\end{figure}

In Figure~\ref{fig:aspl_perf_tree_ecd_relavant_RandomForest}, the orange line represents the average performance of the Mean Encoder across various ASPL values, while the shaded region surrounding the line indicates the 95\% confidence interval. Additionally, we provide the performance of the best encoder responsible for generating the data, serving as a reference for the optimal achievable performance by any encoder.  As observed, the performance of the Mean Encoder drops as ASPL decreases. In fact, the performance deteriorates rapidly as ASPL falls below 25. Conversely, when ASPL is sufficiently large, for instance, 60 or more, the Mean encoder performs very well. Moreover, the performance difference between the Mean Encoder and the best encoder gets smaller and smaller, approaching zero as ASPL increases. 

\begin{figure}[htbp]
	\centering
	\includegraphics[width=0.8\linewidth]{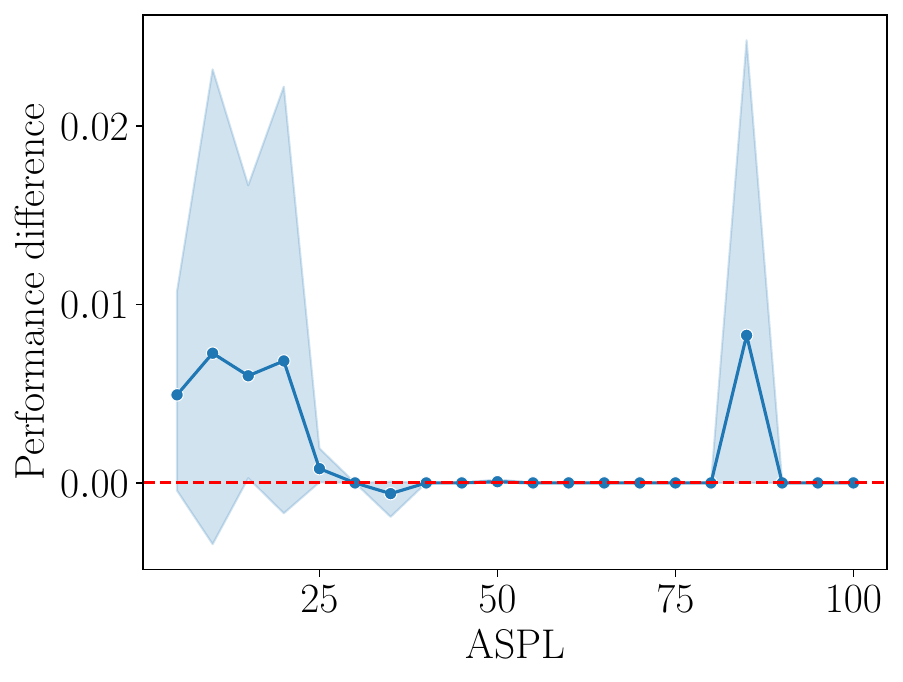}
	\caption{The performance difference, i.e., the accuracy of the SShrink encoder minus the accuracy of the Mean Encoder, on datasets generated from Equation~\eqref{equ:syn_data2} with varying ASPL. The shaded region corresponds to a 95\% confidence interval.
	}
	\label{fig:mean_target_acc_diff}
\end{figure}

The experiment is replicated for the SShrink Encoder, a variant of the Mean Encoder, specifically designed for robust estimation of the conditional mean in scenarios where data is insufficient \cite{micci-barreca2001preprocessing}. As shown in Figure~\ref{fig:mean_target_acc_diff}, SShrink outperforms the Mean Encoder when ASPL is less than 25, although with a marginal advantage of approximately 0.005 accuracy.

%

\section{Quantitative experiment} \label{sec:experiment}

\subsection{Experiment setting}

To validate our findings, we conducted comprehensive experiments on 28 datasets employing various encoders and assessing their performance across common machine learning algorithms. 


The datasets are downloaded from various machine learning repositories and previous studies \cite{pargent2022regularized, valdez-valenzuela2021measuring, seger2018investigation,potdar2017comparative,cerda2018similarity, cerda2022encoding, prokhorenkova2018catboost}. These datasets representing various classification and regression scenarios. Table~\ref{tab:datasets} provides the summary of all datasets. Datasets are sorted by \emph{minASPL}, representing the minimum ASPL across all categorical variables in that dataset. This makes sense because when the variable with the minASPL is data-sufficient, it implies that all categorical variables are data-sufficient. Each dataset is partitioned into training and test sets, comprising 80\% and 20\% of the samples, respectively. Subsequently, we estimated the parameters required for each preprocessing step using the training set and applied them to the test set to avoid information leakage. Standard preprocessing procedures include: filling missing values, encoding all categorical variables with a given encoder and normalization. Further details can be found in the Appendix~\ref{apd:preprocess}. 

Table~\ref{tab:encoders} shows the commonly used encoders categorized by the information they provide. The column ``dimension" denotes the dimensions resulting from the mapping process performed by an encoder. More detailed descriptions of these encoders can be found in the Appendix~\ref{apd:encoder}.

We conducted our experiments with a variety of ATI and Tree-based models implemented by scikit-learn\cite{pedregosa2011scikitlearn}, XGBoost\cite{chen2016xgboost}, and LightGBM\cite{ke2017lightgbm} packages. The ATI models include:
\begin{itemize}
    \item LNR: Linear Regression with L2 pernalty. 
    \item LGR: Logistic Regression with L2 penalty. 
    \item NN: A fully connected neural network with one hidden layer of 100 units and ReLU as the activation function. 
    \item SVM: Support Vector Machine with linear kernel. 
\end{itemize}
The Tree-based models include:
\begin{itemize}
    \item DT: A decision tree with a maximum height of 10 requires at least 10 samples to split an internal node. 
    \item RF: A random forest with 100 decision trees.
    \item XGBoost: A gradient boosting tree model consists of 100 weaker tree estimators.     
    \item LightGBM: A fast gradient boosting tree model consists of 100 weaker tree estimators. 
\end{itemize}

For all models, we utilize the default parameters provided by the packages \cite{pedregosa2011scikitlearn, chen2016xgboost, ke2017lightgbm} due to their broad applicability across various scenarios, offering a good starting point. Extensive parameter tuning can be computationally demanding and may not necessarily improve performance significantly. However, in cases where default settings are inappropriate, we rely on prior experience with similar problems to guide our parameter choices. Further details about the specific settings can be found in the Appendix~\ref{apd:model}.

\begin{table*}[htbp]
	\centering
	\caption{Dataset summary after simple cleanup for both classification and regression tasks.}
	\label{tab:datasets}
		\begin{threeparttable}
			\begin{tabular}{@{}llrrrrrr@{}}
				\toprule
                \multicolumn{1}{c}{\textbf{Task type}} &
				\multicolumn{1}{c}{\textbf{Dataset}} &
				\multicolumn{1}{c}{\textbf{\#Samples}} &
				\multicolumn{1}{c}{\textbf{\#N.Var.\tnote{1}}} &
				\multicolumn{1}{c}{\textbf{\#C.Var.\tnote{2}}} &
				\multicolumn{1}{c}{\textbf{Max Card.}\tnote{3}} &
				\multicolumn{1}{c}{\textbf{PRatio}\tnote{4}} &
				\multicolumn{1}{c}{\textbf{minASPL}\tnote{5}}  \\
				\midrule
				    \multirow{15}[2]{*}{Classification} & Obesity & 2,111  & 3     & 8     & 810     & 0.46  & 3  \\
          & EmployeeAccess & 32,769 & 0     & 9     & 7,518   & 0.94  & 4  \\
          & TripAdvisor & 504   & 4     & 11    & 48    &  0.45  & 11  \\
          & Autism & 704   & 1     & 16    & 67      & 0.27  & 11  \\
          & Kick  & 72,983 & 14    & 14    & 1,063    & 0.12  & 69  \\
          & Churn & 5,000  & 15    & 3     & 51        & 0.14  & 98  \\
          & GermanCredit & 1,000  & 7     & 11    & 10       & 0.70  & 100  \\
          & Mammographic & 961   & 1     & 4     & 8        & 0.46  & 120  \\
          & Wholesale & 440   & 6     & 1     & 3         & 0.32  & 147  \\
          & RoadSafety & 117,536 & 9     & 15    & 380   & 0.21  & 309  \\
          & HIV   & 6,590  & 0     & 8     & 20       & 0.21  & 330  \\
          & CarEvaluation & 1,728  & 0     & 6     & 4      & 0.30  & 432  \\
          & Mushroom & 8,124  & 0     & 18    & 12      & 0.52  & 677  \\
          & Adult & 48,842 & 6     & 8     & 42      & 0.24  & 1,163  \\
          & Nursery & 12,960 & 0     & 8     & 5         & 0.67  & 2,592  \\
    \midrule
    \multirow{13}[1]{*}{Regression} & Colleges & 5935  & 18    & 9     & 2,217   &       & 3  \\
          & CPMP2015 & 2,108  & 22    & 3     & 527     &       & 4  \\
          & EmployeeSalaries & 9,228  & 3     & 4     & 694     &       & 13  \\
          & Moneyball & 1,232  & 6     & 4     & 39       &       & 32  \\
          & Socmob & 1,156  & 1     & 4     & 17     &       & 68  \\
          & Cholesterol & 303   & 6     & 7     & 4         &       & 76  \\
          & StudentPerformance & 1,044  & 15    & 17    & 5       &       & 209  \\
          & Avocado & 18,249 & 11    & 2     & 54       &       & 338  \\
          & BikeSharing & 17,379 & 5     & 6     & 24       &       & 724  \\
          & HousingPrice & 20,640 & 8     & 1     & 5       &       & 4,128  \\
          & Diamonds & 53,940 & 6     & 3     & 8        &       & 6,743  \\
          & CPS1988 & 28,155 & 2     & 2     & 4        &       & 7,039  \\
          & UkAir & 394,299 & 2     & 5     & 53     &       & 7,440  \\
				\bottomrule
			\end{tabular}%
			
			\begin{tablenotes}
                \item [1] The count of total numerical variables. 
                \item [2] The count of total categorical variables. 
				\item [3] The maximum cardinality among all categorical variables.
				\item [4] The ratio of positive samples to the total population in classification tasks. 
				\item [5] The minimum ASPL across all categorical variables in a dataset. It can be obtained by dividing the number of samples by the maximum cardinality.
			\end{tablenotes}
			
	\end{threeparttable}
\end{table*}

\begin{table}[htbp]
	\centering
	\caption{Encoders categorized by the information they provide.}
	\label{tab:encoders}
		\begin{threeparttable}
\begin{tabular}{@{}clr@{}}
	\toprule
	 \textbf{Information}& \textbf{Encoder} & \textbf{Dimension} \\ \midrule
	\multirow{5}{*}{Grouping} & OneHot & $c$ \\
	& BackDiff \cite{anonym2011library} & $c-1$ \\
	& Helmert \cite{anonym2011library} & $c-1$ \\
	& Sum \cite{anonym2011library} & $c-1$ \\
	& BaseN \cite{seger2018investigation} & $\left \lceil \log_b(c+1) \right \rceil$ \\
	\midrule
	\multirow{2}{*}{Ordering} & Ordinal & $1$ \\
	& Count \cite{prokopev2018mean} & $1$ \\
	\midrule
	\multirow{2}{*}{Semantic} & Similarity \cite{cerda2018similarity} & $c$ \\
	& MinHash \cite{cerda2022encoding} & {custom\tnote{1}} \\
	\midrule
	\multirow{5}{*}{Target}
	& Mean & $1$ \\
	& SShrink \cite{micci-barreca2001preprocessing} & $1$ \\
	& MEstimate \cite{micci-barreca2001preprocessing} & $1$ \\
	& JamesStein \cite{morris1983parametric, zhou2015shrinkage, romeijn2017stein} & $1$ \\
	& GLMM \cite{pargent2022regularized} & $1$ \\
	\bottomrule
\end{tabular}%
    \begin{tablenotes}
    \item [1] The default dimension is 30 but can be customized by users. 
    \end{tablenotes}	
			
	\end{threeparttable}
	
\end{table}


The experiments are performed for each dataset-encoder-model combination and are repeated 10 times using different random seeds. Performance measures on the test set are recorded. Specifically, for classification tasks, the F1 score is recorded, defined as $2 * \text{precision} * \text{recall} / (\text{precision} + \text{recall})$, where precision denotes the proportion of samples predicted as positive that are actually positive, and recall represents the proportion of actual positive samples correctly identified as positive. Although accuracy is a straightforward metric for classification, real-world datasets often exhibit class imbalances (as shown in the columm ``PRatio" in Table~\ref{tab:datasets}). Therefore, F1 score is a more reliable performance indicator. In regression tasks, Root Mean Squared Error (RMSE) is used, which indicates the square root of average squared differences between the predicted and actual values. Additionally, for both tasks, the total time (in seconds) used for categorical variables encoding and the model training time is also recorded. 

When evaluating encoders within a dataset and a specific machine learning model, we rank them based on their performance, considering the top-ranked encoder as the best for that particular model-dataset combination. To address the incomparability of performance measures across diverse datasets and tasks, we introduce a relative performance difference metric. This metric quantifies the difference between an encoder's performance and that of the best encoder, divided by the performance of the best encoder. For example, in a regression task using the LNR model, the relative performance of the one-hot encoder is calculated as $\lvert \text{MSE}_\text{OH}-\text{MSE}_\text{Best} \rvert/\text{MSE}_\text{Best}$. 



The implementation of encoders and models is sourced from original authors or third-party libraries such as category-encoders\cite{alkharusi2012categorical}, dirty-cat\cite{cerda2018similarity, cerda2022encoding}, scikit-learn\cite{pedregosa2011scikitlearn}, XGBoost\cite{chen2016xgboost}, and LightGBM\cite{ke2017lightgbm}. All experiments were conducted on laptops equipped with an Intel(R) Core(TM) i7-7920HQ CPU @3.10GHz and 32GB RAM. No GPU acceleration was used. The source code and experimental results are available at \url{https://github.com/QiuRunwen/CategoryEncoderComparison}. 

\subsection{Performance of encoders on ATI models}\label{subsec:perf_of_ecd_in_ati}

In Section~\ref{sec:onehot}, we demonstrated that given sufficient data, the one-hot encoder can reproduce any other encoders for ATI models. This was confirmed through an experiment with a synthetic dataset. In this section, we extend this validation to natural datasets. 


As mentioned, the average number of samples per level (ASPL) indicates data sufficiency. A larger ASPL suggests more sufficient data. Therefore, for a dataset with multiple categorical variables, if the ASPL is sufficient for each categorical variable, we consider the dataset as a whole to be sufficient, as indicated by the \textit{minASPL} of that dataset. 

\begin{figure}[htbp]
	\centering
	\includegraphics[width=0.9\linewidth]{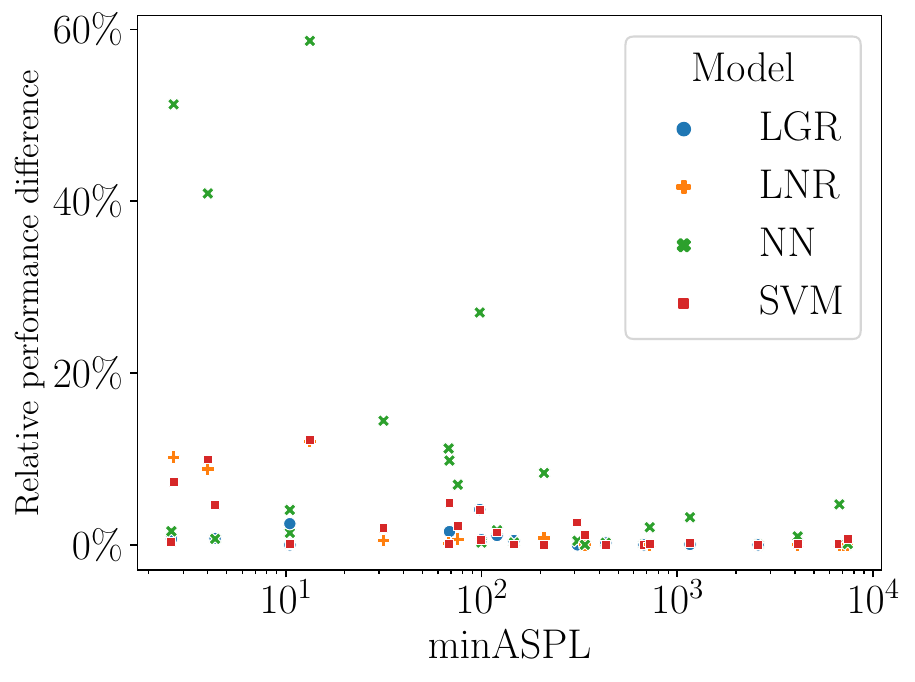}
	\caption{The relative performance difference between the one-hot encoder and the best encoder across all datasets (sorted by minASPL) on four ATI models: LGR, LNR, NN, and SVM. 
}
	\label{fig:performanceofatim}
\end{figure}

Figure~\ref{fig:performanceofatim} illustrates the average relative performance difference between the one-hot encoder and the best encoder across all datasets (sorted by minASPL) on four ATI models: LGR, LNR, NN, and SVM. It can be seen that as minASPL increases, the performance difference gradually diminishes. Specifically, when minASPL is below 100, the one-hot encoder significantly lags behind the best encoder across all ATI models. However, when minASPL is greater or equal 100, the one-hot encoder closely approaches the performance of the best encoder, except for a few scenarios involving NN. Consequently, using a cutoff of 100, we categorize datasets into two groups: sufficient-data (16 datasets) and insufficient-data datasets (12 datasets). 

Next, we investigate the average relative performance difference between the one-hot encoder and the best encoder on all sufficient-data datasets. Table~\subref*{tab:gap_to_best_sf_ati} displays the average and standard deviation of the relative performance difference (\%) between the one-hot encoder and the best encoder across all \textbf{sufficient-data} datasets on four ATI models. Across all four ATI models, the relative performance difference of one-hot encoder is only 0.69\%, indicating a close proximity to the best encoder. More specifically, in the LNR, LGR, and SVM models, the average relative performance difference of the one-hot encoder is less than 0.5\%, while it is 1.06\% in the NN model. An intriguing result is that in LNR and LGR, the one-hot encoder achieves the best average relative performance compared with other encoders. However, in SVM, it lags behind the average best one, the Helmert encoder, by merely 0.13\% and 0.58\% compared with the Sum encoder in NN.

Table~\subref
*{tab:gap_to_best_insf_ati} shows the average and standard deviation of the relative performance difference (\%) across all \textbf{insufficient-data} datasets. We can see that the average relative performance difference of one-hot encoder is 8.82\% on all ATI models. It does not perform well and indicates a considerable deviation from the best encoder. 

Comparing the relative performance difference with other encoders, we observe that the gap between one-hot encoding and the best performer in LNR, LGR, and SVM is 5.41\%, 1.59\%, and 3.96\%, respectively. However, the gap widens significantly in NN, reaching 19.01\%. This substantial difference may be attributed to the complexity of neural network models, and it heavily relies on the sufficiency of the dataset.

Additionally, the top-performing encoders vary under each model. For instance, BackDiff and MinHash performed well in LNR and LGR, but in NN and SVM, they are ranked lower with a substantial relative performance difference. Overall, for ATI models, GLMM ranks in the top when data is insufficient and exhibits relatively stable performance.

\begin{table}[htbp]

	\centering
	\caption{Average and standard deviation of the relative performance difference (\%) between each encoder and the best encoder on four \textbf{ATI models} ($\downarrow$). Bold indicates that an encoder achieves the best results for an ATI model. The underline indicates the one-hot encoder.}
    \label{tab:gap_to_best_ati}

\addtocounter{table}{-1} 
 \subfloat[Over all sufficient-data datasets]{
	\label{tab:gap_to_best_sf_ati}
        \resizebox{\linewidth}{!}{
        \begin{threeparttable}
        \begin{tabular}{@{}lrrrrr@{}}
			\toprule
			\multicolumn{1}{c}{Encoder} &
			\multicolumn{1}{c}{LNR} &
			\multicolumn{1}{c}{LGR} &
			\multicolumn{1}{c}{NN} &
			\multicolumn{1}{c}{SVM} & 
			\multicolumn{1}{c}{Avg.}  \\
			\midrule
     Helmert &  0.12± 0.32 &  0.32± 0.44 &  1.06± 1.62 & \textbf{ 0.32± 0.42} & \textbf{ 0.54± 1.04} \\
    Sum   &  0.12± 0.32 &  0.31± 0.41 & \textbf{ 0.84± 1.51} &  0.72± 1.32 &  0.60± 1.18 \\
    \underline{OneHot} &  \textbf{0.12± 0.30} & \textbf{ 0.27± 0.38} &  1.42± 2.29 &  0.45± 0.72 &  0.69± 1.47 \\
    Similarity & 0.12± 0.30 &  0.32± 0.40 &  1.77± 2.83 &  0.52± 0.87 &  0.84± 1.81 \\
    BackDiff &  0.12± 0.32 &  1.12± 2.34 &  1.84± 2.85 &  1.82± 5.97 &  1.45± 3.91 \\
    MinHash &  1.38± 3.29 &  3.47± 9.52 &  3.64± 5.05 &  5.38±14.22 &  3.86± 9.55 \\
    MEstimate &  2.30± 5.08 &  3.08± 5.59 &  9.51±15.17 &  6.92±21.60 &  6.39±15.46 \\
    SShrink &  2.30± 5.08 &  3.09± 5.63 &  9.54±15.03 &  6.94±21.62 &  6.41±15.43 \\
    Mean  &  2.29± 5.08 &  3.10± 5.64 &  9.65±15.33 &  6.93±21.55 &  6.44±15.50 \\
    GLMM  &  2.29± 5.10 &  4.82±11.16 &  9.63±15.35 &  7.09±22.90 &  6.81±16.54 \\
    JamesStein &  3.20± 5.45 &  3.79± 8.12 & 10.03±16.39 &  7.44±22.25 &  7.00±16.31 \\
    BaseN & 10.79±13.43 & 10.34±19.26 &  4.32± 6.01 & 12.92±24.75 &  9.26±17.51 \\
    Count & 13.48±15.91 & 21.35±24.80 & 21.42±39.72 & 19.90±27.20 & 19.74±29.73 \\
    Ordinal & 14.40±17.78 & 25.30±31.33 & 22.72±42.50 & 22.58±31.46 & 21.94±33.33 \\
			\bottomrule
		\end{tabular}%
    \begin{tablenotes}[para,flushleft]
    Note: Because LNR only runs on regression datasets and LGR only runs on classification datasets, they are weighted when calculating Avg.
    \end{tablenotes}
			
    \end{threeparttable}
        }
    }
    
    \subfloat[Over all insufficient-data datasets]{
    \label{tab:gap_to_best_insf_ati}
    \resizebox{\linewidth}{!}{%
        \begin{threeparttable}
	\begin{tabular}{@{}lrrrrr@{}}
			\toprule
			\multicolumn{1}{c}{Encoder} &
			\multicolumn{1}{c}{LNR} &
			\multicolumn{1}{c}{LGR} &
			\multicolumn{1}{c}{NN} &
			\multicolumn{1}{c}{SVM} & 
			\multicolumn{1}{c}{Avg.}  \\
			\midrule
    GLMM  &  4.68± 5.20 &  2.82± 3.53 & \textbf{ 2.82± 7.34} & \textbf{ 1.09± 1.43} & \textbf{ 2.55± 4.98} \\
    MEstimate &  5.86± 6.44 &  3.05± 4.04 &  3.18± 7.24 &  1.66± 1.75 &  3.10± 5.27 \\
    SShrink &  5.93± 7.23 &  3.13± 4.24 &  3.69± 7.13 &  2.19± 3.02 &  3.47± 5.52 \\
    Mean  &  6.75± 7.25 &  3.07± 4.07 &  4.28± 7.21 &  1.79± 2.11 &  3.66± 5.54 \\
    JamesStein &  7.53± 7.94 &  3.38± 3.82 &  6.09±10.80 &  2.16± 2.53 &  4.57± 7.37 \\
    MinHash &  3.87± 8.01 & \textbf{ 0.21± 0.31} &  9.87± 8.77 &  4.53± 6.43 &  5.48± 7.65 \\
    Similarity &  1.80± 3.82 &  1.77± 1.84 & 11.87±10.48 &  4.63± 6.61 &  6.10± 8.33 \\
    BackDiff & \textbf{ 1.49± 1.45} &  1.62± 2.38 & 13.84±14.75 &  4.76± 6.15 &  6.72±10.45 \\
    BaseN &  7.93±12.16 &  4.32± 8.11 &  9.53±10.02 &  5.82± 9.91 &  7.16± 9.85 \\
    Count & 10.37±13.63 &  7.91± 9.17 &  7.40±12.77 &  8.46±11.98 &  8.33±11.66 \\
    \underline{OneHot} &  5.41± 5.53 &  1.59± 1.50 & 19.01±20.54 &  3.96± 4.07 &  8.82±14.04 \\
    Helmert &  5.10± 5.15 &  1.63± 1.44 & 19.82±21.11 &  4.34± 4.13 &  9.18±14.46 \\
    Sum   &  5.41± 5.56 &  1.59± 1.55 & 48.52±110.73 &  3.96± 4.41 & 18.66±65.76 \\
    Ordinal & 26.26±37.82 & 17.87±26.50 &  8.02±13.57 & 27.40±43.77 & 19.16±32.24 \\
			\bottomrule
		\end{tabular}%
   \begin{tablenotes}[para,flushleft]
    Note: Because LNR only runs on regression datasets and LGR only runs on classification datasets, they are weighted when calculating Avg.
    \end{tablenotes}
			
    \end{threeparttable}
    }
    }
    \addtocounter{table}{1} 
\end{table}

The observed performance of one-hot encoder on both natural sufficient-data and insufficient-data datasets aligns with Theorem~\ref{theorem:onehot_mimic} and the results on synthetic datasets in Section~\ref{sec:onehot}. For sufficient-data datasets, one-hot encoder emerges as the top performer, while for insufficient-data datasets, its performance is constrained by data sufficiency, leading to suboptimal results. Consequently, in scenarios with insufficient data, GLMM is recommended for achieving more robust performance.

\subsection{Performance of encoders on Tree-based models}\label{subsec:perf_of_ecd_in_tb}

In Section~\ref{sec:target}, we argued that target encoders, when provided with sufficient data, can accurately estimate the conditional mean, resulting in excellent performance on Tree-based models. While we initially illustrated this concept using synthetic datasets, our argumentation finds further support in the analysis of natural datasets presented in this section.

\begin{figure}[htbp]
	\centering
	\includegraphics[width=0.9\linewidth]{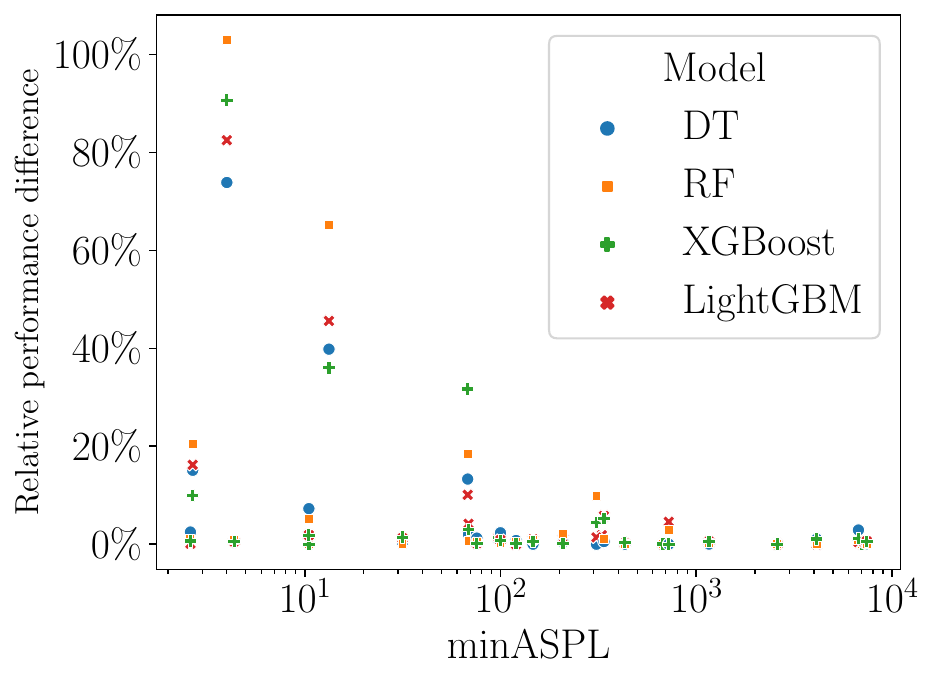}
	\caption{The relative performance difference between the Mean Encoder and the best encoder across all datasets (sorted by minASPL) on four Tree-based models: DT, RF, XGBoost, and LightGBM. }
	\label{fig:performanceoftm}
\end{figure}

In Tree-based models, the criterion for data sufficiency is also the average number of samples per level, consistent with ATI models. Figure~\ref{fig:performanceoftm} illustrates the average relative performance difference between the Mean Encoder and the best encoder across all datasets (sorted by minASPL) for four Tree-based models: DT, RF, XGBoost, and LightGBM. The trend indicates that as minASPL increases, the relative performance difference tends to decrease. The Mean Encoder performs well when minASPL is greater than 100, but its performance falls significantly short of the best encoder in certain scenarios when minASPL is lower. Similarly, we categorize datasets into two groups: sufficient-data datasets (16 datasets) and insufficient-data datasets (12 datasets) using a cutoff of minASPL equal to 100.

Next, we explore the average relative performance difference (\%) among five target encoders -- GLMM, JamesStein, Mean, SShrink, and MEstimate -- compared to the best encoder in both sufficient-data and insufficient-data datasets across four tree-based models. 

Table~\subref*{tab:gap_to_best_sf_tb} presents the average and standard deviation of the relative performance difference (\%) between each encoder and the best encoder across all \textbf{sufficient-data} datasets for four tree-based models. Overall, target encoders, including GLMM, JamesStein, Mean, SShrink, and MEstimate, consistently rank on top. Their average differences with the best encoder are similar, hovering around 1\%. Specifically, in Decision Trees (DT), the Mean Encoder performs the best. In Random Forest (RF), XGBoost, and LightGBM, while MinHash ranks first, target encoders closely follow with comparable performance, demonstrating a marginal difference.

Table~\subref*{tab:gap_to_best_insf_tb} illustrates the relative performance difference of each encoder on all \textbf{insufficient-data} datasets. In general, target encoders exhibit poor performance, while MinHash, Similarity, Sum, and OneHot consistently rank on the top. It is an intriguing observation because these significantly increases the dimensionality on insufficient-data datasets, many of which have high-cardinality variables. 

In summary, our observations indicate that the performance of target encoders on tree-based models is consistent across both natural and synthetic datasets. When there is sufficient data, all target encoders demonstrate strong performance. However, the effectiveness of these encoders is constrained by the adequacy of the available data. On average, MinHash, Similarity, Sum and OneHot encoders serve as viable alternatives in scenarios where the dataset is insufficient.

\begin{table}[htbp]
	\centering
	\caption{Average and standard deviation of the relative performance difference (\%) between each encoder and the best encoder on four \textbf{Tree-based models} ($\downarrow$). Bold indicates that an encoder achieves the best results for an Tree-based model. The underline indicates the target encoders.}
 \label{tab:gap_to_best_tb}

 \addtocounter{table}{-1} 
 \subfloat[Over all sufficient-data datasets]{
	\label{tab:gap_to_best_sf_tb}
        \resizebox{\linewidth}{!}{
 \begin{tabular}{@{}lrrrrr@{}}
				\toprule
				\multicolumn{1}{c}{Encoder} &
				\multicolumn{1}{c}{DT} &
				\multicolumn{1}{c}{RF} &
				\multicolumn{1}{c}{XGBoost} &
				\multicolumn{1}{c}{LightGBM} & 
				\multicolumn{1}{c}{Avg.}  \\
				\midrule
    \underline{GLMM}  &  0.66± 0.95 &  1.16± 2.49 &  0.89± 1.48 &  1.06± 1.75 & \textbf{ 0.94± 1.73} \\
    \underline{JamesStein} &  0.59± 0.86 &  1.11± 2.21 &  0.98± 1.52 &  1.10± 1.62 &  0.94± 1.60 \\
    MinHash &  1.67± 2.78 & \textbf{ 1.04± 1.51} & \textbf{ 0.30± 0.37} & \textbf{ 0.79± 2.12} &  0.95± 1.93 \\
    \underline{Mean}  & \textbf{ 0.52± 0.88} &  1.18± 2.43 &  0.96± 1.57 &  1.14± 1.68 &  0.95± 1.71 \\
    \underline{SShrink} &  0.72± 1.11 &  1.08± 2.04 &  0.95± 1.54 &  1.06± 1.76 &  0.95± 1.61 \\
    \underline{MEstimate} &  0.63± 0.91 &  1.17± 2.36 &  0.94± 1.48 &  1.16± 1.65 &  0.98± 1.66 \\
    Similarity &  2.53± 3.41 &  2.17± 4.28 &  1.24± 2.88 &  1.94± 5.16 &  1.97± 3.96 \\
    Helmert &  6.74±13.51 &  4.11± 8.14 &  2.48± 4.56 &  2.92± 5.59 &  4.06± 8.63 \\
    Sum   &  8.22±16.89 &  4.35± 8.50 &  2.79± 5.11 &  2.92± 5.97 &  4.57±10.23 \\
    OneHot &  8.25±16.88 &  4.37± 8.90 &  2.79± 5.11 &  2.92± 5.74 &  4.58±10.28 \\
    Ordinal &  7.45±14.08 &  5.16± 8.50 &  2.56± 4.24 &  4.25± 7.31 &  4.86± 9.20 \\
    BackDiff &  7.48±14.09 &  6.42±10.49 &  2.46± 4.15 &  4.29± 7.04 &  5.16± 9.65 \\
    BaseN &  7.60±14.04 &  6.81±11.79 &  4.04± 8.07 &  5.91±10.16 &  6.09±11.04 \\
    Count &  7.35±14.27 &  7.78±16.55 &  6.34±15.34 &  7.08±15.85 &  7.14±15.16 \\
				\bottomrule
			\end{tabular}%
    }
    }
    
    \subfloat[Over all insufficient-data datasets]{
    \label{tab:gap_to_best_insf_tb}
    \resizebox{\linewidth}{!}{%

	\begin{tabular}{@{}lrrrrr@{}}
				\toprule
				\multicolumn{1}{c}{Encoder} &
				\multicolumn{1}{c}{DT} &
				\multicolumn{1}{c}{RF} &
				\multicolumn{1}{c}{XGBoost} &
				\multicolumn{1}{c}{LightGBM} & 
				\multicolumn{1}{c}{Avg.}  \\
				\midrule
        MinHash & \textbf{ 2.24± 2.26} &  3.25± 6.18 &  2.33± 2.90 & \textbf{ 1.05± 1.47} & \textbf{ 2.22± 3.64} \\
    Similarity &  2.59± 4.32 &  3.97± 3.85 &  2.10± 2.91 &  1.39± 2.08 &  2.51± 3.43 \\
    Sum   &  4.37± 6.78 & \textbf{ 2.57± 3.31} &  1.37± 1.82 &  2.60± 3.95 &  2.73± 4.35 \\
    OneHot &  4.36± 6.83 &  2.81± 3.37 & \textbf{ 1.35± 1.80} &  2.55± 3.91 &  2.77± 4.37 \\
    Helmert &  2.58± 3.23 &  5.47± 6.25 &  2.48± 3.22 &  2.35± 2.87 &  3.22± 4.21 \\
    Ordinal &  3.67± 4.31 &  3.98± 5.03 &  3.31± 5.36 &  3.02± 3.78 &  3.50± 4.52 \\
    Count &  3.07± 3.87 &  5.07± 9.67 &  5.43±12.69 &  3.88± 7.27 &  4.36± 8.74 \\
    BaseN &  3.82± 4.26 &  4.90± 7.56 &  5.24± 9.24 &  4.15± 5.95 &  4.53± 6.80 \\
    BackDiff &  3.49± 4.18 &  9.60±12.59 &  3.43± 5.34 &  3.01± 3.82 &  4.88± 7.67 \\
    \underline{SShrink} &  8.66±17.43 &  9.84±19.83 &  9.64±18.31 &  8.78±20.84 &  9.23±18.53 \\
    \underline{GLMM}  &  9.26±13.36 & 12.61±19.58 &  9.57±14.58 &  9.75±16.81 & 10.30±15.79 \\
    \underline{MEstimate} & 11.00±17.35 & 13.92±23.10 & 11.25±17.58 & 11.48±19.07 & 11.91±18.82 \\
    \underline{Mean}  & 13.07±22.29 & 18.09±32.64 & 14.78±27.01 & 13.69±25.28 & 14.90±26.26 \\
    \underline{JamesStein} & 14.29±23.64 & 19.08±34.65 & 14.79±27.02 & 13.09±25.02 & 15.31±27.10 \\
				\bottomrule
			\end{tabular}%
        }
    }
    \addtocounter{table}{1} 
\end{table}

\subsection{Impact of encoding on time cost}


In practical scenarios, it becomes critical to take into account the time cost, particularly in large datasets where models exhibit high time complexity, such as Neural Networks (NN), Support Vector Machines (SVM), Random Forests (RF), and XGBoost. In this section, we explore the time costs associated with various encoders when applied to typical large datasets. 


Considering that the number of samples and dimensionality significantly influence time costs, we have selected two datasets for analysis. The first dataset, UkAir, represents a scenario with the largest number of samples. The second dataset, EmployeeAccess, has been chosen due to its high dimensionality after encoding. This selection enables us to examine the time costs associated with each encoder in scenarios where dimensionality plays a crucial role.

\begin{figure}[htbp]
    \centering
    \subfloat[UkAir(\#Samples=394,299, \#C.Var.=5, Max Card.=53)]{
    \includegraphics[width=\linewidth]{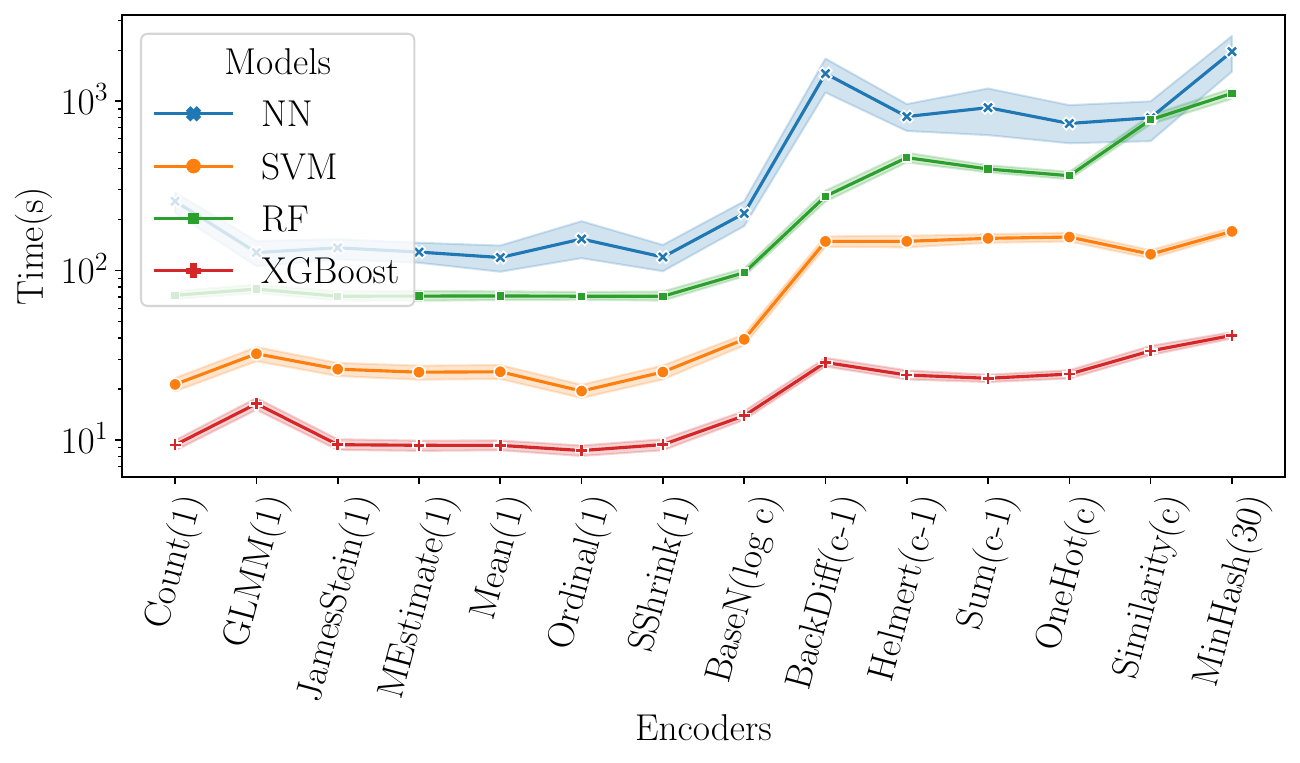}
    \label{fig:time_UkAir_NN_SVM_RF_XGBoost}
    }

    \subfloat[EmployeeAccess(\#Samples=32,769, \#C.Var.=9, Max Card.=7518)]{
        \includegraphics[width=\linewidth]{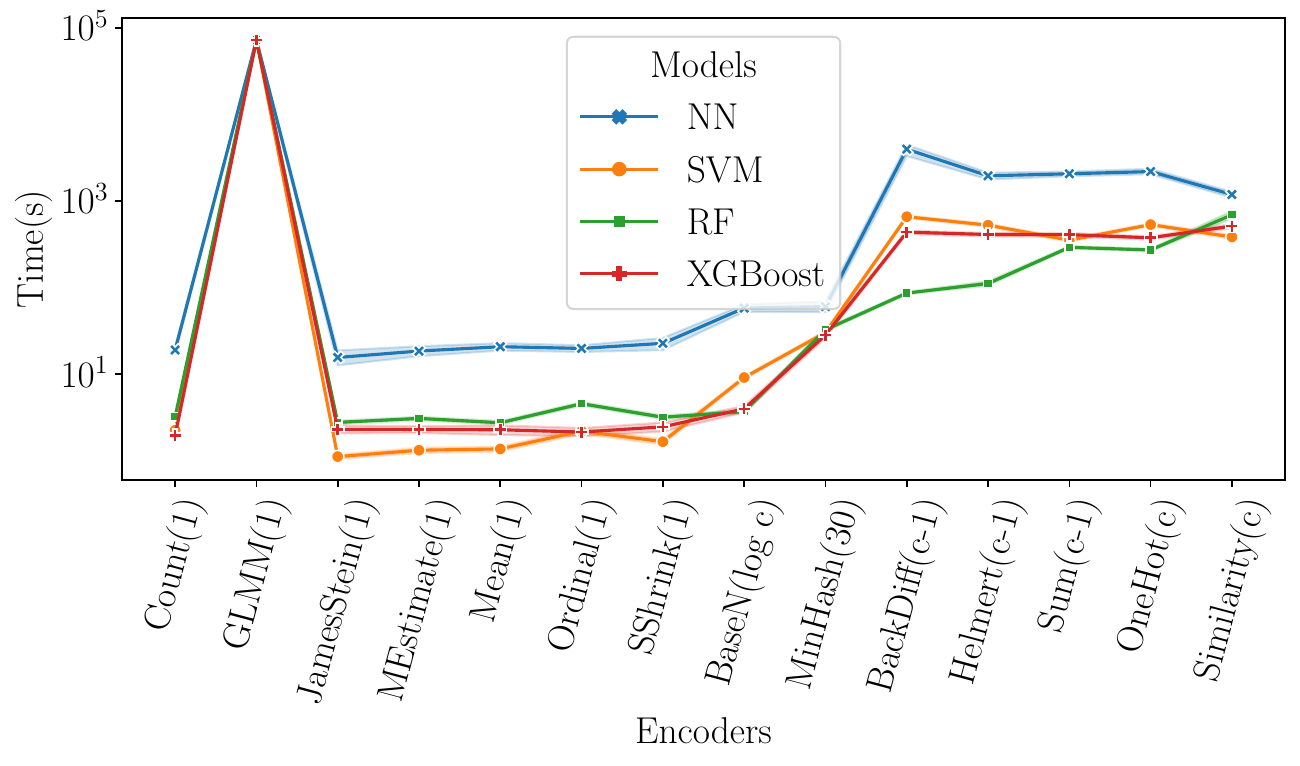}
        \label{fig:time_EmployeeAccess_NN_SVM_RF_XGBoost}
    }
    \caption{Time required for different encoders across various datasets and models. Each subfigure corresponds to a dataset, with each line representing a distinct model. The shaded region corresponds to the 95\% confidence interval. Encoders are sorted based on the dimensionality of the dataset after encoding. The values in parentheses, following each encoder, denote the additional dimensions introduced when dealing with a categorical variable with a cardinality of $c$, as specified in the table~\ref{tab:encoders}. The time cost is the sum of both encoding and model training time.}
    \label{fig:encoder_time_lineplot}
\end{figure}

Figure~\subref*{fig:time_UkAir_NN_SVM_RF_XGBoost} shows the time costs associated with the UkAir dataset, which consists of a total of 394,299 samples. Additionally, this dataset includes 5 categorical variables with a maximum cardinality of 53. On the other hand, Figure~\subref*{fig:time_EmployeeAccess_NN_SVM_RF_XGBoost} shows the time cost required for the EmployeeAccess dataset, comprising a total of 32,769 samples. In this dataset, there are 9 categorical variables and the maximum cardinality reaches 7,518. The trend observed here indicates that the time cost associated with various encoders is primarily influenced by the dimensions resulting from the mapping process performed by an encoder. In other words, it is the dimensions introduced during the encoding of categorical variables that play a crucial role in determining the overall time cost. As the dimensions added increase, the total time required also tends to escalate, with the exception of GLMM. The reason is that GLMM involves fitting multiple generalized linear regression models with various variants. This process becomes more time-consuming, especially when dealing with high-cardinality datasets. 

While we recommend the one-hot encoder for ATI models to achieve good performance when data is sufficient, it does come with a time cost burden due to the introduction of additional $c$ dimensions for each categorical variable. Therefore, in practical scenarios where the time cost is a crucial consideration, the MEstimate encoder offers a good balance between performance and time cost for both sufficient and insufficient datasets in the context of ATI models.

For tree-based models, we argue that Target Encoders should be the top priority, especially when the data is sufficient. This recommendation still holds even when considering the time cost (except GLMM encoder), as these encoders map each categorical variable into a single dimension, making them efficient choices for tree-based models. However, in the case of an insufficient-data dataset, despite MinHash demonstrating the best empirical performance, it encodes each categorical variable into 30 dimensions, introducing additional dimensions to the dataset. Therefore, when time cost becomes a significant factor, Ordinal achieves a favorable trade-off between performance and time cost.


% ...


\section{Practical guide for encoder selection}\label{sec:guide}

\begin{figure}[htbp]
	\centering
	\includegraphics[width=\linewidth]{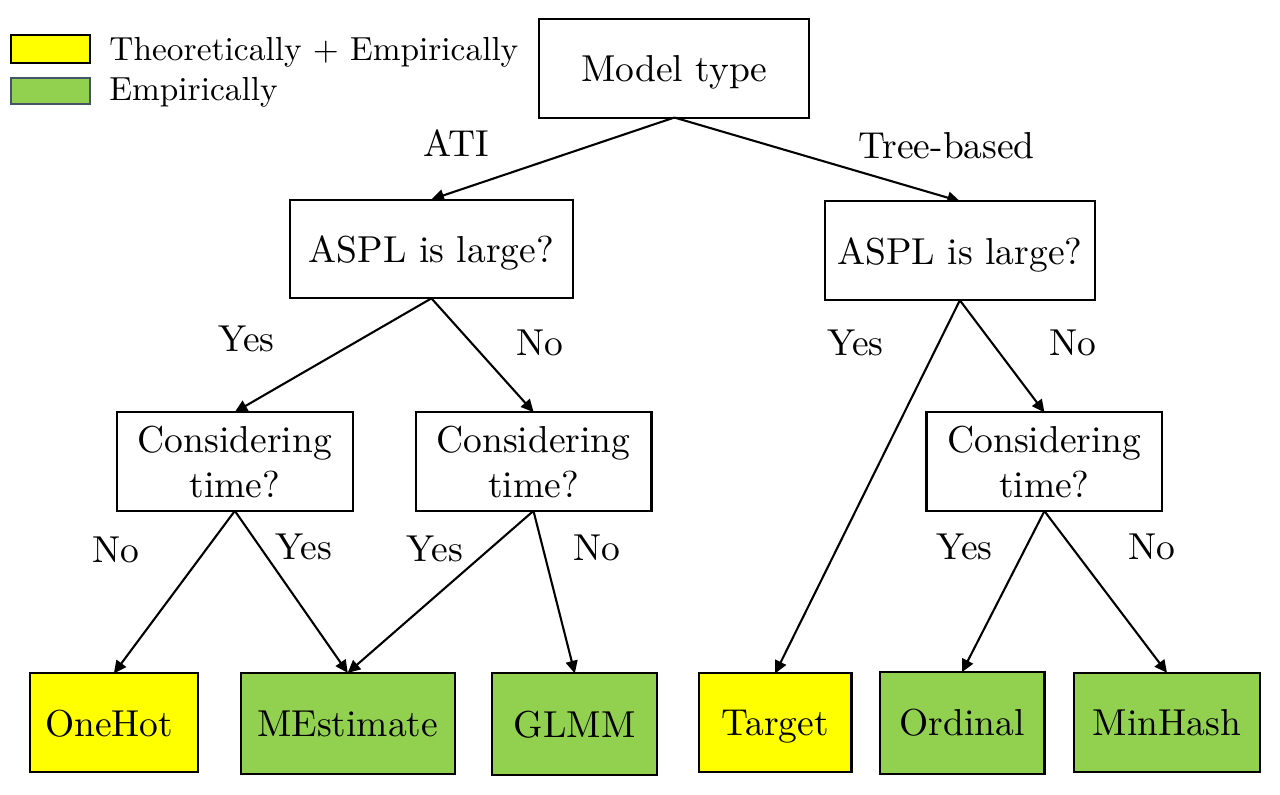}
	\caption{Complete guidance.}
	\label{fig:complete_guidance}
\end{figure}

Based on the theoretical analysis in Section~\ref{sec:onehot} and Section~\ref{sec:target} as well as observations in Section~\ref{sec:experiment}, we present a comprehensive guide for machine learning modelers when encountering datasets with categorical variables. The guide is shown in Figure~\ref{fig:complete_guidance}. 

Firstly, the choice of model type plays a crucial role. For ATI models such as linear regression, logistic regression, neural networks, and support vector machines with a linear kernel, choosing the one-hot encoder is recommended, especially when the average number of samples per level (ASPL) is large.  This recommendation is supported both theoretically and empirically. However, when dealing with large datasets, and time cost is a critical factor, trying the MEstimate encoder is a worthwhile empirical choice. In scenarios where ASPL is small, GLMM serves as an initial encoder worth exploring with average high performance, empirically. It's important to note, however, that GLMM can be time-consuming when encoding high cardinality datasets. If time cost is a significant factor, the MEstimate encoder becomes a worthwhile alternative to consider.

For tree-based models such as random forest, XGBoost, and LightGBM, target encoders exhibit superior performance, supported by both theoretical understanding and empirical experiments. This makes them the preferred choice, especially when the average number of samples per level (ASPL) is substantial. In cases where ASPL is small,  considering MinHash as an initial choice is recommended. 
In situations where ASPL is small, considering MinHash as an initial choice is recommended. If the dataset is extensive, and the time complexity of the model is high, Ordinal becomes a preferable choice when ASPL is small. However, Target Encoders (excluding GLMM) remain a superior choice when ASPL is large.

\section{Conclusion} \label{sec:conclusion}


Categorical variables often appear in datasets for
classification and regression tasks, and they need to be encoded into numerical values prior to training. The choice of encoders to achieve high performance depends on the models while training. In this study, we categorized encoders based on the information they provided during training, and we broadly
classify machine learning models into three categories: 1) ATI models that implicitly perform affine transformations on inputs; 2) Tree-based models that are based on decision trees and 3) others. Theoretically, we prove that the one-hot encoder is the best choice for ATI models in the sense that it can mimic any other encoders by learning suitable weights
from the data. We also explain why the target encoder and its variants are the most suitable encoders for tree-based models. Empirically, we conducted numerous comparative experiments, utilizing 28 datasets (including 15 binary classification tasks and 13 regression tasks), 8 commonly used machine learning models, and 14 encoders to evaluate the performance of encoders, and the result aligns with the theoretical analysis. With all the findings, we provided practical guidelines for modelers to select appropriate encoders when encountering datasets with categorical variables.

In this study, we employ the same encoder for encoding all categorical variables. In future work, we plan to explore the correlation between categorical variables and experiment with combinatorial encoding for all categorical variables. Additionally, while our experiment includes a diverse range of datasets, encoders, and models, certain machine learning models, such as K-Nearest Neighbors and Naive Bayes, have not been included. We plan to address it in our future research.

\ifCLASSOPTIONcompsoc
\section*{Acknowledgments}
\else
\section*{Acknowledgment}
\fi

This work was supported by the National Natural Science Foundation of China (Grant No. U1901222). 


\bibliographystyle{IEEEtranN}

\begin{thebibliography}{29}
	\providecommand{\natexlab}[1]{#1}
	\providecommand{\url}[1]{#1}
	\csname url@samestyle\endcsname
	\providecommand{\newblock}{\relax}
	\providecommand{\bibinfo}[2]{#2}
	\providecommand{\BIBentrySTDinterwordspacing}{\spaceskip=0pt\relax}
	\providecommand{\BIBentryALTinterwordstretchfactor}{4}
	\providecommand{\BIBentryALTinterwordspacing}{\spaceskip=\fontdimen2\font plus
		\BIBentryALTinterwordstretchfactor\fontdimen3\font minus
		\fontdimen4\font\relax}
	\providecommand{\BIBforeignlanguage}[2]{{%
			\expandafter\ifx\csname l@#1\endcsname\relax
			\typeout{** WARNING: IEEEtranN.bst: No hyphenation pattern has been}%
			\typeout{** loaded for the language `#1'. Using the pattern for}%
			\typeout{** the default language instead.}%
			\else
			\language=\csname l@#1\endcsname
			\fi
			#2}}
	\providecommand{\BIBdecl}{\relax}
	\BIBdecl
	
	\bibitem[Cerda and Varoquaux(2022)]{cerda2022encoding}
	P.~Cerda and G.~Varoquaux, ``Encoding {{High-Cardinality String Categorical
			Variables}},'' \emph{IEEE Transactions on Knowledge and Data Engineering},
	vol.~34, no.~3, pp. 1164--1176, Mar. 2022.
	
	\bibitem[Anonym(2011)]{anonym2011library}
	Anonym, ``R {{Library Contrast Coding Systems}} for categorical variables,''
	https://stats.oarc.ucla.edu/r/library/r-library-contrast-coding-systems-for-categorical-variables/,
	Jan. 2011.
	
	\bibitem[Seger(2018)]{seger2018investigation}
	C.~Seger, ``An investigation of categorical variable encoding techniques in
	machine learning: Binary versus one-hot and feature hashing,'' 2018.
	
	\bibitem[Prokopev(2018)]{prokopev2018mean}
	V.~Prokopev, ``Mean (likelihood) encodings: A comprehensive study,''
	https://kaggle.com/code/vprokopev/mean-likelihood-encodings-a-comprehensive-study,
	2018.
	
	\bibitem[Cerda et~al.(2018)Cerda, Varoquaux, and K{\'e}gl]{cerda2018similarity}
	P.~Cerda, G.~Varoquaux, and B.~K{\'e}gl, ``Similarity encoding for learning
	with dirty categorical variables,'' \emph{Machine Learning}, vol. 107, no.~8,
	pp. 1477--1494, Sep. 2018.
	
	\bibitem[Weinberger et~al.(2009)Weinberger, Dasgupta, Langford, Smola, and
	Attenberg]{weinberger2009feature}
	K.~Weinberger, A.~Dasgupta, J.~Langford, A.~Smola, and J.~Attenberg, ``Feature
	hashing for large scale multitask learning,'' in \emph{Proceedings of the
		26th {{Annual International Conference}} on {{Machine Learning}}}, ser.
	{{ICML}} '09.\hskip 1em plus 0.5em minus 0.4em\relax {New York, NY, USA}:
	{Association for Computing Machinery}, Jun. 2009, pp. 1113--1120.
	
	\bibitem[James and Stein(1992)]{james1992estimation}
	W.~James and C.~Stein, ``Estimation with {{Quadratic Loss}},'' in
	\emph{Breakthroughs in {{Statistics}}: {{Foundations}} and {{Basic Theory}}},
	ser. Springer {{Series}} in {{Statistics}}, S.~Kotz and N.~L. Johnson,
	Eds.\hskip 1em plus 0.5em minus 0.4em\relax {New York, NY}: {Springer}, 1992,
	pp. 443--460.
	
	\bibitem[{Micci-Barreca}(2001)]{micci-barreca2001preprocessing}
	D.~{Micci-Barreca}, ``A preprocessing scheme for high-cardinality categorical
	attributes in classification and prediction problems,'' \emph{ACM SIGKDD
		Explorations Newsletter}, vol.~3, no.~1, pp. 27--32, Jul. 2001.
	
	\bibitem[Morris(1983)]{morris1983parametric}
	C.~N. Morris, ``Parametric {{Empirical Bayes Inference}}: {{Theory}} and
	{{Applications}},'' \emph{Journal of the American Statistical Association},
	vol.~78, no. 381, pp. 47--55, Mar. 1983.
	
	\bibitem[Mougan et~al.(2021)Mougan, Masip, Nin, and Pujol]{mougan2021quantile}
	C.~Mougan, D.~Masip, J.~Nin, and O.~Pujol, ``Quantile {{Encoder}}: {{Tackling
			High Cardinality Categorical Features}} in {{Regression Problems}},'' in
	\emph{Modeling {{Decisions}} for {{Artificial Intelligence}}}, ser. Lecture
	{{Notes}} in {{Computer Science}}, V.~Torra and Y.~Narukawa, Eds.\hskip 1em
	plus 0.5em minus 0.4em\relax {Cham}: {Springer International Publishing},
	2021, pp. 168--180.
	
	\bibitem[Romeijn(2017)]{romeijn2017stein}
	J.-W. Romeijn, ``Stein's paradox and group rationality,'' {Faculty of
		Philosophy University of Groningen}, {ColaForm Workshop Paris 2017}, Tech.
	Rep., 2017.
	
	\bibitem[Zhou(2015)]{zhou2015shrinkage}
	X.~Zhou, ``Shrinkage {{Estimation}} of {{Log-odds Ratios}} for {{Comparing
			Mobility Tables}},'' \emph{Sociological Methodology}, vol.~45, no.~1, pp.
	320--356, Aug. 2015.
	
	\bibitem[Alkharusi(2012)]{alkharusi2012categorical}
	H.~Alkharusi, ``Categorical {{Variables}} in {{Regression Analysis}}: {{A
			Comparison}} of {{Dummy}} and {{Effect Coding}},'' \emph{International
		Journal of Education}, vol.~4, no.~2, p. 202, Jun. 2012.
	
	\bibitem[Moeyersoms and Martens(2015)]{moeyersoms2015including}
	J.~Moeyersoms and D.~Martens, ``Including high-cardinality attributes in
	predictive models: {{A}} case study in churn prediction in the energy
	sector,'' \emph{Decision Support Systems}, vol.~72, pp. 72--81, Apr. 2015.
	
	\bibitem[Potdar et~al.(2017)Potdar, S., and D.]{potdar2017comparative}
	K.~Potdar, T.~S., and C.~D., ``A {{Comparative Study}} of {{Categorical
			Variable Encoding Techniques}} for {{Neural Network Classifiers}},''
	\emph{International Journal of Computer Applications}, vol. 175, no.~4, pp.
	7--9, Oct. 2017.
	
	\bibitem[Wright and K{\"o}nig(2019)]{wright2019splitting}
	M.~N. Wright and I.~R. K{\"o}nig, ``Splitting on categorical predictors in
	random forests,'' \emph{PeerJ}, vol.~7, p. e6339, Feb. 2019.
	
	\bibitem[Hancock and Khoshgoftaar(2020)]{hancock2020survey}
	J.~T. Hancock and T.~M. Khoshgoftaar, ``Survey on categorical data for neural
	networks,'' \emph{Journal of Big Data}, vol.~7, no.~1, p.~28, Apr. 2020.
	
	\bibitem[Prokhorenkova et~al.(2018)Prokhorenkova, Gusev, Vorobev, Dorogush, and
	Gulin]{prokhorenkova2018catboost}
	L.~Prokhorenkova, G.~Gusev, A.~Vorobev, A.~V. Dorogush, and A.~Gulin,
	``{{CatBoost}}: Unbiased boosting with categorical features,'' in
	\emph{Advances in {{Neural Information Processing Systems}}}, vol.~31.\hskip
	1em plus 0.5em minus 0.4em\relax {Curran Associates, Inc.}, 2018.
	
	\bibitem[Johnson and Khoshgoftaar(2021)]{johnson2021encoding}
	J.~M. Johnson and T.~M. Khoshgoftaar, ``Encoding {{Techniques}} for
	{{High-Cardinality Features}} and {{Ensemble Learners}},'' in \emph{2021
		{{IEEE}} 22nd {{International Conference}} on {{Information Reuse}} and
		{{Integration}} for {{Data Science}} ({{IRI}})}, 2021, pp. 355--361.
	
	\bibitem[{Valdez-Valenzuela} et~al.(2021){Valdez-Valenzuela}, {Kuri-Morales},
	and {Gomez-Adorno}]{valdez-valenzuela2021measuring}
	E.~{Valdez-Valenzuela}, A.~{Kuri-Morales}, and H.~{Gomez-Adorno}, ``Measuring
	the {{Effect}} of {{Categorical Encoders}} in {{Machine Learning Tasks Using
			Synthetic Data}},'' in \emph{Advances in {{Computational Intelligence}}},
	ser. Lecture {{Notes}} in {{Computer Science}}, I.~Batyrshin, A.~Gelbukh, and
	G.~Sidorov, Eds.\hskip 1em plus 0.5em minus 0.4em\relax {Cham}: {Springer
		International Publishing}, 2021, pp. 92--107.
	
	\bibitem[Seca and {Mendes-Moreira}(2021)]{seca2021benchmark}
	D.~Seca and J.~{Mendes-Moreira}, ``Benchmark of {{Encoders}} of {{Nominal
			Features}} for {{Regression}},'' in \emph{Trends and {{Applications}} in
		{{Information Systems}} and {{Technologies}}}, ser. Advances in {{Intelligent
			Systems}} and {{Computing}}, {\'A}.~Rocha, H.~Adeli, G.~Dzemyda, F.~Moreira,
	and A.~M. Ramalho~Correia, Eds.\hskip 1em plus 0.5em minus 0.4em\relax
	{Cham}: {Springer International Publishing}, 2021, pp. 146--155.
	
	\bibitem[Pargent et~al.(2022)Pargent, Pfisterer, Thomas, and
	Bischl]{pargent2022regularized}
	F.~Pargent, F.~Pfisterer, J.~Thomas, and B.~Bischl, ``Regularized target
	encoding outperforms traditional methods in supervised machine learning with
	high cardinality features,'' \emph{Computational Statistics}, Mar. 2022.
	
	\bibitem[Fisher(1958)]{fisher1958grouping}
	W.~D. Fisher, ``On {{Grouping}} for {{Maximum Homogeneity}},'' \emph{Journal of
		the American Statistical Association}, vol.~53, no. 284, pp. 789--798, Dec.
	1958.
	
	\bibitem[Breiman et~al.(1984)Breiman, Friedman, Olshen, and
	Stone]{breiman1984classification}
	L.~Breiman, J.~H. Friedman, R.~A. Olshen, and C.~J. Stone, \emph{Classification
		{{And Regression Trees}}}, 1st~ed.\hskip 1em plus 0.5em minus 0.4em\relax
	{Routledge}, 1984.
	
	\bibitem[Pedregosa et~al.(2011)Pedregosa, Varoquaux, Gramfort, Michel, Thirion,
	Grisel, Blondel, Prettenhofer, Weiss, and Dubourg]{pedregosa2011scikitlearn}
	F.~Pedregosa, G.~Varoquaux, A.~Gramfort, V.~Michel, B.~Thirion, O.~Grisel,
	M.~Blondel, P.~Prettenhofer, R.~Weiss, and V.~Dubourg, ``Scikit-learn:
	{{Machine}} learning in {{Python}},'' \emph{the Journal of machine Learning
		research}, vol.~12, pp. 2825--2830, 2011.
	
	\bibitem[Chen and Guestrin(2016)]{chen2016xgboost}
	T.~Chen and C.~Guestrin, ``{{XGBoost}}: {{A Scalable Tree Boosting System}},''
	in \emph{Proceedings of the 22nd {{ACM SIGKDD International Conference}} on
		{{Knowledge Discovery}} and {{Data Mining}}}, ser. {{KDD}} '16.\hskip 1em
	plus 0.5em minus 0.4em\relax {New York, NY, USA}: {Association for Computing
		Machinery}, Aug. 2016, pp. 785--794.
	
	\bibitem[Ke et~al.(2017)Ke, Meng, Finley, Wang, Chen, Ma, Ye, and
	Liu]{ke2017lightgbm}
	G.~Ke, Q.~Meng, T.~Finley, T.~Wang, W.~Chen, W.~Ma, Q.~Ye, and T.-Y. Liu,
	``{{LightGBM}}: {{A Highly Efficient Gradient Boosting Decision Tree}},'' in
	\emph{Advances in {{Neural Information Processing Systems}}}, vol.~30.\hskip
	1em plus 0.5em minus 0.4em\relax {Curran Associates, Inc.}, 2017.
	
	\bibitem[Zhang(2015)]{zhang2015tips}
	O.~Zhang, ``Tips for data science competitions,'' Jul. 2015.
	
	\bibitem[{Deepanshu Bhalla}(2015)]{deepanshubhalla2015weight}
	{Deepanshu Bhalla}, ``Weight of {{Evidence}} ({{WOE}}) and {{Information
			Value}} ({{IV}}) {{Explained}},'' Mar. 2015.
	
\end{thebibliography}

\clearpage
{\appendices
\section{Details of datasets}\label{apd:dataset}
\subsection{Datasets for Classification}


\textbf{Obesity}. This dataset include data for the estimation of obesity levels in individuals from the countries of Mexico, Peru and Colombia, based on their eating habits and physical condition. Downloaded from \url{https://archive.ics.uci.edu/dataset/544/estimation+of+obesity+levels+based+on+eating+habits+and+physical+condition}.


\textbf{EmployeeAccess}. The data consists of real historical data collected from 2010 \& 2011.
Employees are manually allowed or denied access to resources over time.
The data is used to create an algorithm capable of learning from this historical data
to predict approval/denial for an unseen set of employees. Downloaded from \url{https://www.kaggle.com/competitions/amazon-employee-access-challenge}.

\textbf{TripAdvisor}. This dataset includes quantitative and categorical features from online reviews from 21 hotels located in Las Vegas Strip, extracted from TripAdvisor (http://www.tripadvisor.com). The target is to predict the scores of the hotels. Downloaded from \url{https://archive.ics.uci.edu/dataset/397/las+vegas+strip}.

\textbf{Autism}. Autistic Spectrum Disorder Screening Data for Adult. The target is to predict whether a person has Autistic Spectrum Disorder. Downloaded from \url{https://archive.ics.uci.edu/dataset/426/autism+screening+adult}.

\textbf{Kick}. One of the biggest challenges of an auto dealership purchasing a used car at an auto auction is the risk of that the vehicle might have serious issues that prevent it from being sold to customers. The auto community calls these unfortunate purchases "kicks". The challenge of this competition is to predict if the car purchased at the Auction is a Kick (bad buy). Downloaded from \url{https://www.openml.org/d/41162}.

\textbf{Churn}. A dataset relating characteristics of telephony account features and usage and whether or not the customer churned. Downloaded from \url{https://www.openml.org/d/41283}.

\textbf{GermanCredit}. This dataset classifies people described by a set of attributes as good or bad credit risks.
700 good and 300 bad credits with 20 predictor variables. Data from 1973 to 1975. 
Stratified sample from actual credits with bad credits heavily oversampled. Downloaded from \url{https://archive.ics.uci.edu/dataset/522/south+german+credit}.

\textbf{Mammographic}. Discrimination of benign and malignant mammographic masses based on BI-RADS attributes and the patient's age. Downloaded from \url{https://archive.ics.uci.edu/dataset/161/mammographic+mass}.

\textbf{Wholesale}. The data set refers to clients of a wholesale distributor. It includes the annual spending in monetary units (m.u.) on diverse product categories. The target is to predict whether the sale channel is Horeca (Hotel/Restaurant/CafÃ©) or Retail channel (Nominal). Downloaded from \url{https://archive.ics.uci.edu/dataset/292/wholesale+customers}.

\textbf{RoadSafety}. It is the traffic accidents data from United Kingdom, downloaded from \url{https://data.gov.uk/dataset/cb7ae6f0-4be6-4935-9277-47e5ce24a11f/road-safety-data}. Predicting whether ``Accident\_Severity'' is fatal or serious. Extracted month from ``Date'' as we expect it is related to weather, which may have an important impact on traffic accidents. We also extracted hours from ``Time'', as traffic conditions vary greatly at different hours of the day. We dropped ``1st\_Road\_Number'' and ``2nd\_Road\_Number'' as there are too many distinct values.

\textbf{HIV}. The data contains lists of octamers (8 amino acids) and a flag (-1 or 1) depending on whether HIV-1 protease will cleave in the central position (between amino acids 4 and 5). Downloaded from \url{https://archive.ics.uci.edu/dataset/330/hiv+1+protease+cleavage}.

\textbf{CarEvaluation}. It was derived from a simple hierarchical decision model originally developed for the demonstration of a Expert system for decision making. The model proposed should be able to evaluates cars according to concept structure, i.e. multiple categorical features. Downloaded from \url{http://archive.ics.uci.edu/ml/datasets/Car+Evaluation}.

\textbf{Mushroom}. From Audobon Society Field Guide; mushrooms described in terms of physical characteristics; classification: poisonous or edible. Downloaded from \url{https://archive.ics.uci.edu/dataset/73/mushroom}.

\textbf{Adult}. It is extracted by Barry Becker from the 1994 Census database. The target is to predict whether income exceeds \$50K/yr based on census data. Also known as "Census Income" dataset. \url{https://archive.ics.uci.edu/dataset/2/adult}.

\textbf{Nursery}. It  was derived from a hierarchical decision model
originally developed to rank applications for nursery schools. It was used during several years in 1980's when there was excessive enrollment to these schools in Ljubljana, Slovenia, and the rejected applications
frequently needed an objective explanation. The final decision depended on three subproblems: occupation of parents and child's nursery, family structure and financial standing,
and social and health picture of the family. The model proposed should be able to predict whether the nursery is recommended for the child or not. Downloaded from \url{https://archive.ics.uci.edu/dataset/76/nursery}.

\subsection{Datasets for Regression}
\textbf{Colleges}. It contains information about U.S. colleges and schools downloaded from \url{https://beachpartyserver.azurewebsites.net/VueBigData/DataFiles/Colleges.txt}. Predicting ``Mean Earning 6 Years''. Columns ``Median Earning 6 Years'', ``Mean Earning 10 Years'' and ``Median Earning 10 Years'' are removed as they correlate highly to the prediction target.

\textbf{CPMP2015}. It is a benchmark result of the Container Pre-Marshalling Problem (CPMP).
The task is to predict the runtime of a CPMP algorithm on a given instance. Downloaded from \url{https://www.openml.org/d/41700}.

\textbf{EmployeeSalaries}. It includes annual salary information for 2016 for Montgomery County, Maryland employees, downloaded from \url{https://catalog.data.gov/dataset/employee-salaries-2016}. The target it to predict the ``Current Annual Salary''.  The ``Date First Hired'' is converted to the number of days to the latest date as a proxy to measure how long the employee has worked. The ``Department'' column is dropped as it contains abbreviations of ``Department Name''.

\textbf{Moneyball}. It is gathered from baseball-reference.com. The original author wants to know which important factors affect the performance of the baseball team . In our experiment, the target is to predict Runs Scored (RS). Downloaded from \url{https://www.openml.org/d/41021}.

\textbf{Socmob}. This dataset described social mobility, i.e. how the sons' occupations are related to their fathers' jobs.
An instance represent the number of sons that have a certain job A given the father has the job B (additionally conditioned on race and family structure). The dataset was originally collected for the survey of "Occupational Change in a Generation II". The version 
we use here is the one from OpenML, which is a preprocessed version of the original dataset. The target is to predict the number of people when given the information of these occupation information. Downloaded from \url{https://www.openml.org/d/44987}.

\textbf{Cholesterol}. It contains 4 databases concerning heart disease diagnosis. The earliest one is from UCI \url{https://archive.ics.uci.edu/dataset/45/heart+disease}. In OpenML, the dataset is called cholesterol. The target variable is ``chol'' (serum cholestoral in mg/dl), which is a regression task. Downloaded from \url{https://www.openml.org/d/204}.

\textbf{StudentPerformance}. This data approach student achievement in secondary education of two Portuguese schools. The data attributes include student grades, demographic, social and school related features) and it was collected by using school reports and questionnaires. Two datasets are provided regarding the performance in two distinct subjects: Mathematics (mat) and Portuguese language (por). The target is to predict the final grade ``G3''. Downloaded from \url{https://archive.ics.uci.edu/dataset/320/student+performance}.

\textbf{Avocado}. It is a historical dataset on avocado prices and sales volume in multiple US markets. It is originally from the Hass Avocado Board website in May of 2018. We downloaded the dataset from Kaggle \url{https://www.kaggle.com/neuromusic/avocado-prices}. The target is to predict the ``AveragePrice'' of the avocado.

\textbf{HousingPrice}. It is about the median house prices for California districts derived from the 1990 census. It is a famous regression dataset used in many textbooks, such as ``Hands-On Machine learning with Scikit-Learn and TensorFlow''. The target is to predict the ``median\_house\_value''. Downloaded from \url{https://www.kaggle.com/datasets/camnugent/california-housing-prices}.

\textbf{BikeSharing}. This dataset contains the hourly and daily count of rental bikes between years 2011 and 2012 in Capital bikeshare system with the corresponding weather and seasonal information. The target is to predict the count of total rental bikes including both casual and registered. Downloaded from \url{https://archive.ics.uci.edu/dataset/275/bike+sharing+dataset}.

\textbf{Diamonds}. This classic dataset contains the prices and other attributes of almost 54,000 diamonds. The target is to predict the prices of diamonds. Downloaded from \url{https://www.openml.org/d/44979}.

\textbf{CPS1988}. Cross-section data originating from the March 1988 Current Population Survey by the US Census Bureau. The data is a sample of men aged 18 to 70 with positive annual income greater than USD 50 in 1992, who are not self-employed nor working without pay. The target is to predict the wage (in dollars per week). Downloaded from \url{https://www.openml.org/d/43963}.

\textbf{UkAir}. Hourly particulate matter air polution data of Great Britain for the year 2017, provided by Ricardo Energy and Environment on behalf of the UK Department for Environment, Food and Rural Affairs (DEFRA) and the Devolved Administrations. The target is to predict the PM2.5 hourly measured. Downloaded from \url{https://www.openml.org/d/42207}.

\section{Data preprocess procedure}\label{apd:preprocess}

For original datasets, the following cleanup steps are performed:
\begin{itemize}
    \item { Simple Conversion}: Convert special values such as ``N.A" to missing; convert the string representation of a date and time value to its DateTime equivalent.
    \item { Simple Feature Extraction}: Convert date to the number of days since a reference date; convert datetime to time passed since a reference time; extract year, month, day, day-of-week, week-of-year.
    \item { Drop Useless Columns}: Columns with no obvious meaning; textual columns; columns with unique values such as ID; columns with more than 50\% missing values; columns with a single value; columns with fewer than two samples per value.
\end{itemize}

After the cleanup steps, we partitioned the dataset into training and test sets, comprising 80\% and 20\% of the samples, respectively. Subsequently, we estimated the parameters required for each preprocessing step using the training set and applied them to the test set to avoid information leakage.

Firstly, we filled in the missing values in the datasets. Specifically, missing values in numerical variables were replaced with the average value of that variable within the training set. Categorical variables with missing values were replaced with the most frequent category observed within the training set. 

Subsequently, all categorical variables were transformed into numerical ones using the same encoding method.

Finally, normalization was applied to each variable within the training set. This involved standardizing each feature by removing the mean and scaling to unit variance.


\section{Details of encoders}\label{apd:encoder}
\subsection{Grouping Encoders}
Grouping Encoders focuses on grouping information. Its characteristic is that encoding is a bijection. We can deduce the original value from the encoded value. Another characteristic is that the encoded dimension is often larger than 1.

\textbf{One-hot Encoder}. It is the most widely used encoding method. Feature with the cardinality of $c$ will be encoded into $c$ variables. Only variables at the corresponding level are 1, and other variables are 0 (i.e. $\begin{pmatrix}0  &\dots & 0 & 0 & 1 \end{pmatrix}^T$ indicates the first level,  $\begin{pmatrix}0  &\dots & 0 & 1 & 0 \end{pmatrix}^T$ indicates the second level, indicates the third level of $\begin{pmatrix}0 &\dots & 1 & 0 & 0 \end{pmatrix}^T$ , and so on.).

\textbf{BaseN Encoders} \footnote{\url{https://contrib.scikit-learn.org/category_encoders/basen.html} }. It uses the system of numeration to encode the categorical features. For example, when the base is 2, it becomes binary encoding ($\begin{pmatrix}0 &\dots & 0 & 0 & 1 \end{pmatrix}^T$ indicates the first level,  $\begin{pmatrix}0 &\dots & 0 & 1 & 0 \end{pmatrix}^T$ indicates the second level, indicates the third level of $\begin{pmatrix}0 &\dots & 0 & 1 & 1 \end{pmatrix}^T$ , and so on.). \cite{seger2018investigation} indicates that the performance of binary encoding will be slightly worse than that of one-hot, but the storage will be much less, especially when using neural networks. Note that sometimes the performance of binary encoding will be far worse than one-hot encoding. We set `base=2', which means binary encoder is applied in the experiment.

\textbf{Contrast Encoders}\cite{anonym2011library}. It is usually used in statistics. Most of them are used in scenarios that need explanation. For example, in linear regression, a feature with the cardinality of $c$ will be encoded into $c-1$ variables, so that the parameters multiplied by variables can represent the contrast relationship between some levels. For example, suppose we are using the season to predict temperature. SP, SU, FA, and WI are encoded into 3 variables respectively by dummy coding. Then each parameter represents the temperature that how the much higher current season is than the default season, which shows in Equation~\ref{equ:dunmmy encoding}. 
\begin{equation} \label{equ:dunmmy encoding}
	y=b+\underbrace{w_1}_{\text{SP vs. WI}} x_1+\underbrace{w_2}_{\text{SU vs. WI}} x_2+\underbrace{w_3}_{\text{FA vs. WI}} x_3
\end{equation}
Other contrast encoding methods are similar. \textbf{Backward Difference Encoder} makes the coefficient indicate the contrast between the current level and the next level. \textbf{Helmert Encoder} makes the coefficient indicate the contrast between the current level and the levels all behind. \textbf{Sum Encoder} makes the coefficient indicate the contrast between the current level and the average of all levels. It is worth noting that they are less frequently used in machine learning than one-hot encoder.

\subsection{Ordering Encoders}
Ordering encoding focuses on ordering information, which believes that there is an order between levels in the categorical feature.

\textbf{Ordinal Encoder}. It is a very common encoding that encodes each level in order. The ordinal encoding can reach good results when the nature of the categorical feature is ordinal and the order is known\cite{valdez-valenzuela2021measuring}. However, because it is cumbersome and difficult to assign the order manually, we usually use the order in which they appear for encoding. For example, the first level appearing in the training set is encoded as 1, the second level is encoded as 2, and the $c^{\text{th}}$ level is encoded as $c$.

\textbf{Count Encoder}. It encodes the categorical features by counting. For example, if ``Male" appears 100 times in the training set, the ``Male" is encoded as 100. In a comparative experiment in Kaggle\cite{prokopev2018mean}, the count encoding works well on many datasets when using the gradient boosting tree classifier/regressor.

\subsection{Semantic Encoders}
Semantic encoding focuses on semantic information. It mainly encodes some long texts, such as shopping reviews, case descriptions, etc. This kind of categorical feature usually has a very high cardinality, and the average sample in each cell is small. In addition, long texts usually contain emotional information, semantic similarity, context information, etc. which we uniformly call semantic information.


\textbf{Similarity Encoder}\cite{cerda2018similarity} and \textbf{min-hash Encoder}\cite{cerda2022encoding} 
do not need to train a model for embedding. Their basic idea is that the char or string is the basic unit of information. The higher the overlap of char or sub-string at different levels, the closer their values should be after encoding. For example, in the position title, the overlap between ``police officer" and ``police aide" is higher and more relevant, while the correlation with "mechanical technician" is weak. Specifically, e.g., 3-grams(Paris) = \{Par, ari, ris\} and 3-grams(Parisian) = \{Par, ari, ris, isi, sia, ian\} have three 3-grams in common, and their similarity is $sim_\text{3-gram}$(Paris, Parisian) = 3. The advantage of this method is that it can capture the semantic information of categorical features without additional corpus and models. But string similarity is not always equivalent to semantic similarity, and its encoded value may bring noise. Therefore, this encoding method is more suitable for the automatic processing of non-curated strings, e.g. the subjective answers to the questionnaire.

In \textbf{Similarity Encoder}, we set the main parameters as \textit{ngram\_range=(2,4)}, which means it will extract multiple character strings with lengths of 2, 3, and 4 in order from the string. They are all default paramerters provided by the authors.

In \textbf{MinHash Encoder}, And we set the main parameters as \textit{n\_components=30, ngram\_range=(2,4)}, which means the output dimension of a categorical variable is 30. They are all default paramerters provided by the authors.

\subsection{Target Encoders (conditional mean)} \label{subsec:TargetEncodersCondMean}
Target encoding focuses on target information. Its characteristic is that it utilizes the information in the target variable $y$. In other words, it is supervised. In most cases, they take the conditional expectation (or conditional probability) as the encoded value. binary classification has the same encoding formula as regression, when the target variable is preprocessed to only have 0 and 1. Another characteristic is that 1 categorical feature is usually encoded as 1 dimension variable, which makes it popular when the dataset is large.


Bayesian encoding\cite{micci-barreca2001preprocessing} is an Empirical Bayes method. It mainly combines prior distribution and posterior distribution to estimate the expected value of the dependent variable. The level $v_k$ is encoded as $\mathbb{E} (y|x=v_k)$ .The general formula used in Empirical Bayes estimation is Equation~\eqref{equ:Empirical Bayes estimation}.
\begin{equation} \label{equ:Empirical Bayes estimation}
    \phi(v_k)=B_k\hat{\mu_k}+(1-B_k)\mu
\end{equation}
where
\begin{itemize}
    \item $\mu$ is the prior value of $y$, which is the estimated value of $\mathbb{E}(y)$. It is the average value of $y$ in the training set because the number of instances in the training set is usually large enough.
    \item $\hat{\mu}_k$ is posterior value of $y$,which is the estimated value of $\mathbb{E}(y|x=v_k)$.  It is usually the average value of $y|x=v_k$. Note that it is unreliable when the number of instances $m_k$ for $x=v_k$ is small.
    \item $B_k$ is \textit{shrinkage factor}, which measures the weight of prior value and posterior value ($0 \le B_ k\le 1$). If it is more confident to estimate the posterior value accurately, $B_k$ should be more closed to 1, otherwise, it tends to be 0.
\end{itemize}
Different $\mu$, $B_k$, and $\hat{\mu}_k$ bring about different Bayesian encoding methods.

\textbf{Mean Encoder}. The naive idea: use the mean to estimate the conditional expectation. If expressed by Equation~\eqref{equ:Empirical Bayes estimation}, mean encoding only considers the posterior value, that is $B_k=1$. The posterior value is calculated as the target mean at the corresponding level. Supposed the dataset satisfied $x^{(i)}=v_k$ is $\mathcal{D}_k$, and $m_k$ is the size of it. Then mean encoding of the level $v_k$ can be expressed as:
\begin{equation}  \label{equ:mean encoding}
    \phi(v_k)=\frac{1}{m_k}\sum_{i\in \mathcal{D}_k} y_i
\end{equation}

\textbf{S-shrink Encoder}\cite{micci-barreca2001preprocessing}. According to the idea of Equation~\eqref{equ:Empirical Bayes estimation}, the mean of $y$ in the training set is used as the prior value, and the mean of $y|x=v_k$ is used as the posterior value. When the number of instances $m_k$ is small, the mean of $y|x=v_k$ is unreliable. So $B_k$ is set as a monotonic increasing function of $m_k$, which is also an s-shaped function. The corresponding value is shown in Equation~\eqref{equ:s-shrink encoding}
\begin{equation} \label{equ:s-shrink encoding}
    \mu=\frac{1}{m}\sum_{i=1}^my_i ,\
    \hat{\mu}_k=\frac{1}{m_k}\sum_{i\in \mathcal{D}_k}y_i , \
    B_k=\frac{1}{1+\exp(-\frac{m_k-S_1}{S_2})}
\end{equation} 
Where there are two hyperparameters:
\begin{itemize}
    \item $S_1$ is the threshold related to the number of the instance under the level. When $m_k<S_1$, then $B_k<0.5$, which means the prior value is more reliable than posterior value. The value of parameter uses the default value of the library, which is 20.
    \item $S_2$ is the steepness related to the function between $B_k$ and $m_k$. The function is steeper if $S_2$ is smaller. When $S_2\to 0$, then $S_1$ will be the hard threshold, which means $B_k=0$ when $m_k<S_1$. The value of parameter uses the default value of the library, which is 10.
\end{itemize}

\textbf{M-estimate Encoder}\cite{micci-barreca2001preprocessing}. It is also can be expressed by Equation~\eqref{equ:Empirical Bayes estimation}. The corresponding value is shown in Equation~\eqref{equ:m-estimate encoding}. Where $M$ is a hyper-parameter about the threshold. When $m_k<M$, then $B_k<0.5$, which means the prior value is more reliable than the posterior value. The value of parameter $M$ uses the default value of the library, which is 1.
\begin{equation} \label{equ:m-estimate encoding}
    \mu=\frac{1}{m}\sum_{i=1}^my_i ,\
    \hat{\mu}_k=\frac{1}{m_k}\sum_{i\in \mathcal{D}_k}y_i , \
    B_k=\frac{m_k}{m_k+M}
\end{equation}

\textbf{James-Stein Encoder}. This encoding method mainly uses the idea of James-Stein estimation. \citet{james1992estimation} proved that when the target value is the normal distribution and three or more parameters are estimated at the same time (in our case it can be expressed as $c\ge 3$), James–Stein Estimator always achieves lower mean squared error (MSE) than the maximum likelihood estimator. That is to say, it is not optimal to estimate the actual mean value with the sample mean value of each level. Different formulas can be derived from different distribution assumptions\cite{morris1983parametric,zhou2015shrinkage,romeijn2017stein}.
The version we used supposes the target is a normal distribution. It is also can be expressed by Equation~\eqref{equ:Empirical Bayes estimation}. The details are as follows.
\begin{equation} \label{equ:James-Stein encoding}
	\mu=\frac{1}{m}\sum_{i=1}^my_i ,\
	\hat{\mu}_k=\frac{1}{m_k}\sum_{i\in \mathcal{D}_k}y_i , \
	1-B_k=\frac{c-3}{c-1}\frac{\hat{\sigma_k^2}}{\hat{\sigma_k^2}+\tau^2}
\end{equation}
The equation is also used in binary classification, although the target variable is definitely not normally distributed as probabilities are bound to lie on interval [0,1]. But according to the documentation\footnote{\url{https://github.com/scikit-learn-contrib/category_encoders/blob/06e46db07ff15c362d7de5ad225bb0bffc245369/category_encoders/james_stein.py\#L290C13-L290C13}} and our practice, this version is better than the version that assumes the target is a binomial distribution.

\textbf{Quantile Encoder}\cite{mougan2021quantile}. It replaces the posterior mean with the value of the posterior quantile in M-estimate. $\mu$ and $\hat\mu_k$is replaced by quantile. The median is usually used as a quantile. The method may work when there are many outliers in the target value. Note that this method applies to the case where the target is continuous. If the target is binary as 0-1, the encoded value can only be 0,1 or a few corresponding quantile values. It is not suitable for binary classification.
Its performance is similar to other target encoders and is not displayed in the paper.

In addition, \textbf{Leave-One-Out Encoder}\cite{zhang2015tips} and \textbf{CatBoost Encoder}\cite{prokhorenkova2018catboost} are not encoders according to our definition of a general encoder which maps each level $x \in \mathcal{V}$ into $l$-dimensional vector $\phi(x)$. Because the encodings of these two encoders are based on each sample in the training set, instead of using level as the encoding unit. Even if they have the same level, their encoding values are different.

\subsection{Target Encoders (other)} \label{subsec:TargetEncodersOther}
\textbf{Weight Of Evidence Encoder} (WOE)\cite{deepanshubhalla2015weight}. Its common version is only applicable to binary classification. Its performance is similar to other target encoders and is not displayed in the paper.

\textbf{Generalized Linear Mixed Model Encoding} (GLMM) \footnote{\url{https://contrib.scikit-learn.org/category_encoders/glmm.html}}. Although it does not take the estimation form of Equation~\eqref{equ:Empirical Bayes estimation}, it's a regularized target encoding\cite{pargent2022regularized}. For the regression, it encodes the categorical variables by one-hot encoder first. Then for each categorical variable, its encoded value is used to train a linear regression with a intercept.
\begin{equation*}
    y=b+\boldsymbol{w}^T\phi_\text{OH}(x)+\epsilon
\end{equation*}
where $\boldsymbol{w}=\left[w_1,w_2,...,w_c\right]^T$. After fitting, the corresponding coefficient of a level is seen as its encoded value, which is also called random effect since the encoding is seen as a random variable $w_k\sim N(0,\tau^2)$. But it is worth mentioning that the fitting method is not the commonly used least squares method, because the basic assumption is that not every sample is independent of each other.
The encoding can be seen as:
\begin{equation*}
    \phi_\text{GLMM}(v_k)= w_k = \mathbb{E}[y] - \mathbb{E}[y \mid x=v_k] \quad \forall k=1,2,\cdots,c
\end{equation*}
For the binary classification, it is almost the same as the regression. The regression equation uses logistic regression, which is also called a generalized linear model.
The details can be seen in the reference\footnote{\url{https://www.statsmodels.org/stable/generated/statsmodels.regression.mixed_linear_model.MixedLM.html}, \url{https://www.statsmodels.org/stable/generated/statsmodels.genmod.bayes_mixed_glm.BinomialBayesMixedGLM.html}}. Because multiple generalized linear mixed models need to be trained when there are several categorical variables, the encoding time is relatively long\cite{seca2021benchmark}.

\section{Details of models}\label{apd:model}
All hyper-parameters of the models are set to their default values as provided by the libraries cited in the references \cite{pedregosa2011scikitlearn, chen2016xgboost, ke2017lightgbm}, except for Decision Tree where pruning will be required to avoid overfitting if the default parameters are applied.

\textbf{Neural Network (NN)}. Multi-layer Perceptron classifier\footnote{\url{https://scikit-learn.org/stable/modules/generated/sklearn.neural_network.MLPClassifier.html}} and Multi-layer Perceptron regressor\footnote{\url{https://scikit-learn.org/stable/modules/generated/sklearn.neural_network.MLPRegressor.html}} are multiple-hidden-layer networks implemented by scikit-learn \cite{pedregosa2011scikitlearn}. The log-loss function is used in classification, while squared error function is used in regression. All parameters use the following default values:
\begin{itemize}
    \item \textit{hidden\_layer\_sizes=(100,)}. One hidden layer of 100 neurons.
    \item \textit{activation=`relu'}. Use ReLU as activation function for the hidden layer.
    \item \textit{solver=`adam'}. The solver for weight optimization. We use Adam, which is a popular optimizer.
    \item \textit{alpha=0.0001}. Strength of the L2 regularization term. The L2 regularization term is divided by the sample size when added to the loss.
    \item  \textit{batch\_size=`auto'}. Size of minibatches for stochastic optimizers. When set to ``auto'', \textit{batch\_size=min(200, n\_samples}).
    \item \textit{learning\_rate\_init=0.001}. The initial learning rate used. It controls the step-size in updating the weights.
    \item  \textit{max\_iter=200}. Maximum number of iterations. The solver iterates until convergence (determined by ``tol'') or this number of iterations. It  is the actually the number of iterations because we do not use early stop.
    \item \textit{shuffle=True}. Whether to shuffle samples in each iteration. 
    \item \textit{beta\_1=0.9}. Exponential decay rate for estimates of first moment vector in adam.
    \item  \textit{beta\_2=0.999}. Exponential decay rate for estimates of second moment vector in adam.
    \item \textit{epsilon=1e-8}. Value for numerical stability in adam. 
\end{itemize}

\textbf{Logistic Regression (LGR)}. Logistic Regression Classifier \footnote{\url{https://scikit-learn.org/stable/modules/generated/sklearn.linear_model.LogisticRegression.html}} is implemented by scikit-learn \cite{pedregosa2011scikitlearn}. All parameters use the following default values:
\begin{itemize}
    \item \textit{penalty=`l2'}. Add a L2 penalty term.
    \item \textit{C=1.0}. Inverse of regularization strength; must be a positive float. Like in support vector machines, smaller values specify stronger regularization.
    \item \textit{fit\_intercept=True}. Specifies if a constant (a.k.a. bias or intercept) should be added to the decision function.
    \item \textit{solver=`lbfgs'}. Algorithm to use in the optimization problem.
    \item \textit{max\_iter=100}. Maximum number of iterations taken for the solvers to converge.
\end{itemize}

\textbf{Linear Regression (LNR)}\footnote{\url{https://scikit-learn.org/stable/modules/generated/sklearn.linear_model.RidgeCV.html}}. Linear least squares with l2 regularization implemented by scikit-learn \cite{pedregosa2011scikitlearn}. All parameters use the following default values:
\begin{itemize}
    \item \textit{fit\_intercept=True}. Whether to calculate the intercept for this model. If set to false, no intercept will be used in calculations (i.e. data is expected to be centered).
    \item \textit{alphas=(0.1, 1.0, 10.0)}. Array of alpha values to try. Regularization strength; must be a positive float. Regularization improves the conditioning of the problem and reduces the variance of the estimates. Larger values specify stronger regularization.
    \item \textit{cv=None}. Use the efficient Leave-One-Out cross-validation.
    \item \textit{gcv\_mode=`auto'}. Flag indicating which strategy to use when performing Leave-One-Out Cross-Validation. `auto' : use `svd' if n\_samples $>$ n\_features, otherwise use `eigen'
`svd' : force use of singular value decomposition of $X$ when $X$ is
    dense, eigenvalue decomposition of $X^TX$ when $X$ is sparse.
'eigen' : force computation via eigenvalue decomposition of $XX^T$.
\end{itemize}

\textbf{Support Vector Machine (SVM)}. Linear Support Vector Classifier\footnote{\url{https://scikit-learn.org/stable/modules/generated/sklearn.svm.LinearSVC.html}} and Linear Support Vector Regressor\footnote{\url{https://scikit-learn.org/stable/modules/generated/sklearn.svm.LinearSVR.html}} are implemented by scikit-learn \cite{pedregosa2011scikitlearn}. All parameters use the following default values:
\begin{itemize}
    \item \textit{penalty=`l2'}. Specifies the norm used in the penalization. It is only used on the classifier and not on the regressor.
    \item \textit{loss=`squared\_hinge'} is for classifier. Specifies the loss function. `squared\_hinge' is the square of the hinge loss. \textit{loss=`epsilon\_insensitive'} is for the regressor. The squared epsilon-insensitive loss (`squared\_epsilon\_insensitive') is the L2 loss.
    \item \textit{epsilon=0.0} is for the regressor, while there is no this parameter for the classifier. Epsilon parameter in the epsilon-insensitive loss function.
    \item \textit{dual=True}. Select the algorithm to either solve the dual or primal optimization problem.
    \item \textit{tol=1e-4}. Tolerance for stopping criteria.
    \item \textit{C=1.0}. Regularization parameter. The strength of the regularization is inversely proportional to $C$.
    \item \textit{fit\_intercept=True}. Fit an intercept. The feature vector is extended to include an intercept term: $[x_1, ..., x_n, 1]$, where $1$ corresponds to the intercept.
    \item \textit{intercept\_scaling=1.0}. The instance vector x becomes $[x_1, ..., x_n, \text{intercept\_scaling}]$, i.e. a ``synthetic" feature with a constant value equal to intercept\_scaling is appended to the instance vector.
    \item \textit{class\_weight=None} is for the classifier, while there is no this parameter for the regressor. No class weight. 
    \item \textit{max\_iter=1000}. The maximum number of iterations to be run.
\end{itemize}

\textbf{Decision Tree (DT)}. The decision tree classifier\footnote{\url{https://scikit-learn.org/stable/modules/generated/sklearn.tree.DecisionTreeClassifier.html}} and decision tree regressor\footnote{\url{https://scikit-learn.org/stable/modules/generated/sklearn.tree.DecisionTreeRegressor.html}} are all implemented by scikit-learn \cite{pedregosa2011scikitlearn}. The \textit{max\_depth} of a tree is 10, and an internal node will be split only if it contains at least \textit{min\_samples\_leaf=10}. All other parameters use the following default values:
\begin{itemize}
    \item \textit{criterion=``gini"} is for the classifier while \textit{criterion=``squared\_error"} is for the regressor. The function to measure the quality of a split. ``gini" is the Gini impurity. ``squared\_error" is the mean squared error, which is equal to variance reduction as feature selection criterion and minimizes the L2 loss using the mean of each terminal node.
    \textit{splitter=``best"}. Always choose the best variable and best threshold to split a node.
    \item \textit{min\_samples\_leaf=1}. The minimum number of samples required to be at a leaf node.
    \item \textit{max\_leaf\_nodes=None}. No limit on the number of leaves.
    \item \textit{ccp\_alpha= 0.0}. No pruning is performed.
\end{itemize}

\textbf{Random Forest (RF)}. The random forest classifier\footnote{\url{https://scikit-learn.org/stable/modules/generated/sklearn.ensemble.RandomForestClassifier.html}} and random forest regressor\footnote{\url{https://scikit-learn.org/stable/modules/generated/sklearn.ensemble.RandomForestRegressor.html}} are implemented by scikit-learn \cite{pedregosa2011scikitlearn}. All parameters use the following default values:
\begin{itemize}
    \item \textit{n\_estimators=100}. The number of trees in the forest.
    \item \textit{criterion=``gini"} is for the classifier while \textit{criterion=``squared\_error"} is for the regressor. The function to measure the quality of a split. ``gini" is the Gini impurity. ``squared\_error" is the mean squared error, which is equal to variance reduction as feature selection criterion and minimizes the L2 loss using the mean of each terminal node.
    \item \textit{max\_depth=None}. The maximum depth of the tree. If None, then nodes are expanded until all leaves are pure or until all leaves contain less than \item \textit{min\_samples\_split} samples.
    \item \textit{min\_samples\_split=2}. The minimum number of samples required to split an internal node.
    \item \textit{splitter=``best"}. Always choose the best variable and best threshold to split a node.
    \item \textit{min\_samples\_leaf=1}. The minimum number of samples required to be at a leaf node.
    \item \textit{max\_features=``sqrt"}. The number of features to consider when looking for the best split, i.e. \textit{max\_features=sqrt(n\_features)}.
    \item \textit{bootstrap=True}. Bootstrap samples are used when building tree.
    \item \textit{max\_leaf\_nodes=None}. No limit on the number of leaves.
    \item \textit{ccp\_alpha= 0.0}. No pruning is performed.
\end{itemize}

\textbf{XGBoost}. XGBClassifier and XGBRegressor are popular models proposed by \citet{chen2016xgboost} which can be seen in \url{https://xgboost.readthedocs.io/en/stable/python/python\_api.html}. All parameters use the following default values\footnote{\url{https://xgboost.readthedocs.io/en/stable/parameter.html}}:
\begin{itemize}
    \item \textit{objective=`binary:logistic'} represents logistic regression for binary classification, while \textit{objective=`reg:squarederror'} represents regression with squared loss.
    \item \textit{eta=0.3}. Alias of \textit{learning\_rate}.Step size shrinkage used in update to prevents overfitting. After each boosting step, we can directly get the weights of new features, and eta shrinks the feature weights to make the boosting process more conservative.
    \item \textit{gamma=0}. Alias of \textit{min\_split\_loss}. Minimum loss reduction required to make a further partition on a leaf node of the tree. 
    \item \textit{max\_depth=6}. Maximum depth of a tree.
    \item \textit{min\_child\_weight=1} Minimum sum of instance weight (hessian) needed in a child. If the tree partition step results in a leaf node with the sum of instance weight less than min\_child\_weight, then the building process will give up further partitioning. In linear regression task, this simply corresponds to minimum number of instances needed to be in each node. The larger min\_child\_weight is, the more conservative the algorithm will be. 
    \item \textit{max\_delta\_step=2}. Maximum delta step we allow each leaf output to be. If the value is set to 0, it means there is no constraint. If it is set to a positive value, it can help making the update step more conservative.
    \item \textit{subsample=1} Subsample ratio of the training instances. We use the full training set instead of subsample set.
    \item \textit{colsample\_by*=1}. \textit{*} can be `node', `tree' or `level' This is a family of parameters for subsampling of columns. Specify the fraction of columns to be subsampled. We do not sample the columns.
    \item \textit{lambda=1}. Alias of \textit{reg\_lambda}. L2 regularization term on weights. 
    \item \textit{alpha=0}. Alias of \textit{reg\_alpha}. L1 regularization term on weights. We use L2 instead of L1.
    \item \textit{tree\_method=`auto'} The tree construction algorithm used in XGBoost. See description in the reference paper and Tree Methods. Faster histogram optimized approximate greedy algorithm.
    \item \textit{scale\_pos\_weight=1} Control the balance of positive and negative weights. 
    \item \textit{max\_bin=256}. Maximum number of discrete bins to bucket continuous features.
\end{itemize}

\textbf{LightGBM}. LGBMClassifier\footnote{\url{https://lightgbm.readthedocs.io/en/latest/pythonapi/lightgbm.LGBMClassifier.html}} and LGBMRegressor\footnote{\url{https://lightgbm.readthedocs.io/en/latest/pythonapi/lightgbm.LGBMRegressor.html}} are popular models proposed by \citet{ke2017lightgbm}. All parameters use the following default values:
\begin{itemize}
    \item \textit{num\_leaves=31}. Maximum tree leaves for base learners.
    \item \textit{max\_depth=-1} Maximum tree depth for base learners, <=0 means no limit.
    \item \textit{learning\_rate=0.1}. Boosting learning rate. 
    \item \textit{n\_estimators=100}. Number of boosted trees to fit.
    \item \textit{subsample\_for\_bin=200000}. Number of samples for constructing bins.
    \item \textit{objective} `regression' for LGBMRegressor, `binary' for LGBMClassifier.
    \item \textit{class\_weight=None}. All classes are supposed to have weight one. 
    \item \textit{min\_split\_gain=0.}. Minimum loss reduction required to make a further partition on a leaf node of the tree.
    \item \textit{min\_child\_weight=1e-3}. Minimum sum of instance weight (Hessian) needed in a child (leaf).
    \item \textit{min\_child\_samples=20}. Minimum number of data needed in a child (leaf).
    \item \textit{subsample=1}. Subsample ratio of the training instance. We do not subsample.
    \item \textit{subsample\_freq=0}. Frequency of subsample, $<=0$ means no enable. We do not subsample.
    \item \textit{colsample\_bytree=1}. Subsample ratio of columns when constructing each tree. We do not subsample.
    \item \textit{reg\_alpha=0}. L1 regularization term on weights.
    \item \textit{reg\_lambda=0}. L2 regularization term on weights.
\end{itemize}

}

\vfill

\end{document}